\documentclass{article}
\usepackage[utf8]{inputenc}

\usepackage{hyperref}       
\usepackage{url}            
\usepackage{booktabs}       
\usepackage{amsfonts}       
\usepackage{nicefrac}       
\usepackage{comment}
\usepackage{mismath} 
\usepackage{wrapfig}
\usepackage{microtype}      
\usepackage{xcolor}         
\usepackage{bbm}
\usepackage{amsmath}
\DeclareMathOperator*{\argmin}{arg\,min}
\DeclareMathOperator*{\argmax}{arg\,max}
\usepackage{amssymb}
\usepackage{algorithm}
\usepackage{algpseudocode}

\usepackage{graphicx}
\graphicspath{{./figures/}}
\usepackage{caption}
\usepackage{subcaption}

\usepackage{acronym}
\acrodef{DMD}[DMD]{\emph{Dynamic Mode Decomposition}}
\acrodef{CMD}[CMD]{\emph{Correlation Mode Decomposition}}
\acrodef{GD}[GD]{\emph{Gradient Descent}}
\acrodef{SGD}[SGD]{\emph{Stochastic Gradient Descent}}
\acrodef{NN}[NN]{\emph{Neural Network}}

\title{The Underlying Correlated Dynamics in Neural Training}
\author{Rotem Turjeman \and Tom Berkov \and Ido Cohen \and Guy Gilboa\\
Technion - Israel Institute of Technology\\
Haifa, Israel\\
{\tt\small \{srotemtu, ptom, idoc\}@campus.technion.ac.il} \\
{\tt\small guy.gilboa@ee.technion.ac.il} 
}
\date{\today}

\begin{document}

\maketitle
\begin{abstract}
Training of neural networks is a computationally intensive task. The significance of understanding and modeling the training dynamics is growing as increasingly larger networks are being trained. We propose in this work a model based on the correlation of the parameters' dynamics, which dramatically reduces the dimensionality. We refer to our algorithm as \emph{correlation mode decomposition} (CMD). It splits the parameter space into groups of parameters (modes) which behave in a highly correlated manner through the epochs. 

We achieve a remarkable dimensionality reduction with this approach, where networks like ResNet-18, transformers and GANs, containing millions of parameters, can be modeled well using just a few modes. We observe each typical time profile of a mode is spread throughout the network in all layers. Moreover, our model induces regularization which yields better generalization capacity on the test set. This representation enhances the understanding of the underlying training dynamics and can pave the way for designing better acceleration techniques. 
\end{abstract}

\section{Introduction}
\label{sec:intro}

In recent years, with the success of large \ac{NN}  architectures featuring numerous parameters, there is an increasing interest in modeling the training  behavior. The aim is to gain better understanding on the very complex process of training and to develop improved methods to save memory or to accelerate training. Concretely, one can view the inference or the training process as flows of data, that is --  dynamical processes generated by ODEs. This viewpoint has been adopted by several studies. In \cite{haber2017stable} the authors apply the stability notion to architectures, making the forward pass stable thus mitigating the vanishing or exploding gradient phenomena. Other works, such as \cite{lawrence2020almost}, discuss stability of network architectures as well. The methods of \cite{chen2018neural}, \cite{NEURIPS2019_21be9a4b} and \cite{NEURIPS2019_01386bd6} offer memory consumption reduction, precision to training time trading and some performance improvements. In \cite{NEURIPS2019_812b4ba2}  adversarial training is accelerated. More research that use either stochastic or deterministic ODE frameworks can be found in \cite{huh2020time}, \cite{chen2020dynamical}, \cite{mignacco2020dynamical}, \cite{mei2018mean} and the references therein. A kernel analysis of the flow and linearization of wide networks has significantly contributed to the modeling and understanding of gradient descent in neural networks \cite{jacot2018neural,lee2019wide}. This, for instance, allowed to estimate well the training time also of standard architecture (non-wide) networks \cite{zancato2020predicting}.

Another related approach is to estimate the gradient descent process by approximating the evolution operator of the latent dynamics and decomposing it into modes. A leading algorithm is \ac{DMD} \cite{schmid2010dynamic}, which is completely data driven and very intuitive to use. \ac{DMD} fits any dynamical process by a linear system. This may result in oversimplifying the dynamics of highly nonlinear systems. We discuss the literature related to \ac{DMD} on networks and provide a short analysis of problems this model might yield in Section \ref{sec:DMD}. 
Our experiments and analysis show that \ac{DMD} and its variants do not model well the highly complex dynamics of \ac{NN} training. 

\begin{figure} [H]
    \centering
    \includegraphics[width=0.9\linewidth]{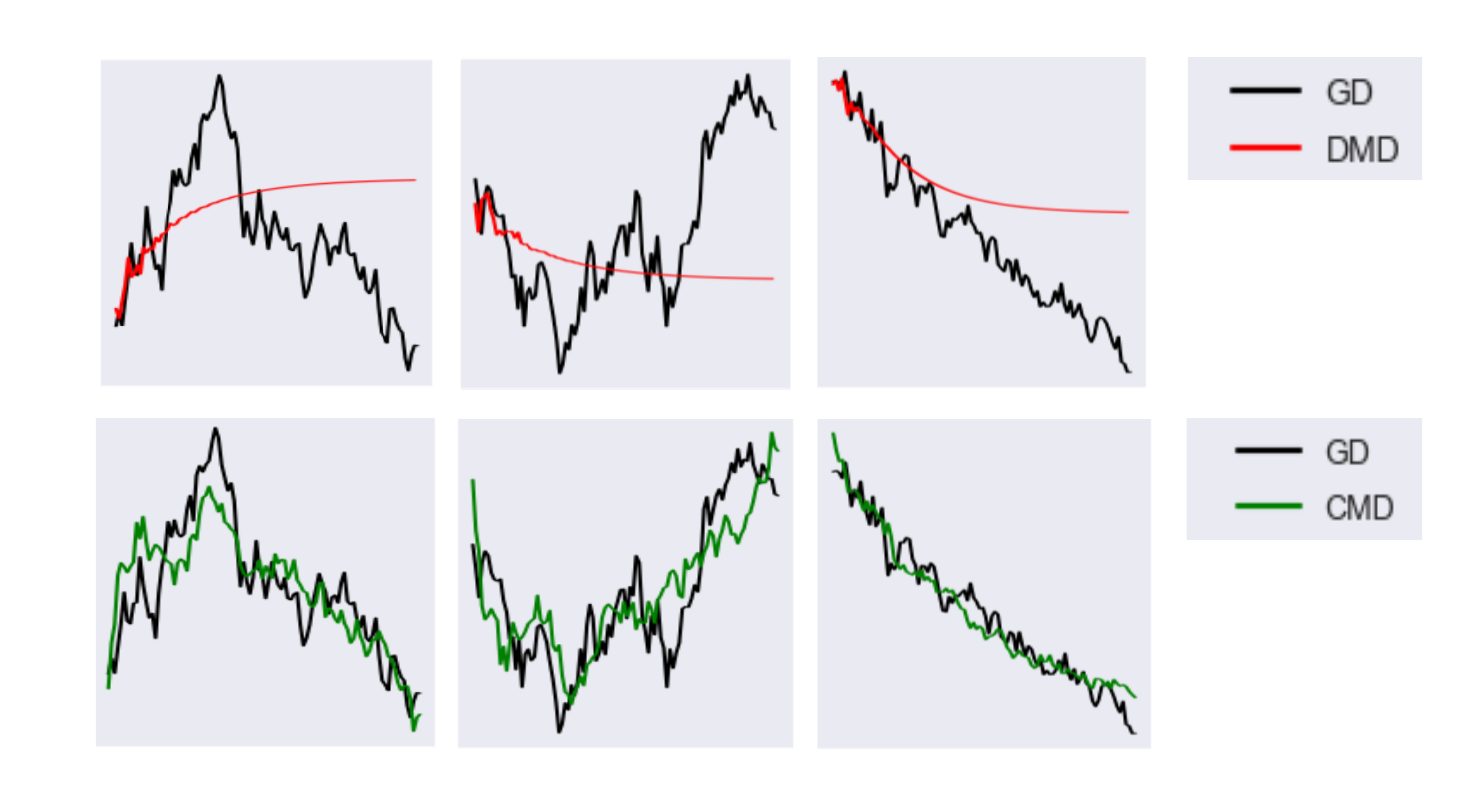}
    \caption{Three sampled weights from CIFAR10 classification with Resnet18 training (GD training in black), and their corresponding reconstruction using DMD modeling (\textbf{Top}) and CMD modeling (\textbf{Bottom}). CMD produces much more faithful approximations.}
    \label{cmd_vs_dmd_rec}
\end{figure}

Our main observation is that while the network parameters may behave non-smoothly in the training process, many of them are highly correlated and can be grouped into "Modes" characterized by their correlation to one common evolution profile. 
We thus propose \acf{CMD}, a simple and efficient method to model the network's dynamics. Let us outline the approximations of DMD and CMD. The training dynamic provided by DMD is in the form of, 
\begin{equation}\label{eq:DMDdyn}
    w_i^k \approx \sum_{j=1}^{r}\alpha_j\phi_j(i)(\lambda_j)^k,
\end{equation}
where $w_i^k$ is the network parameter indexed $i$ at epoch $k$, and $\{\phi, \lambda\}$ are DMD eigenvectors and eigenvalues respectively.
The dynamic model according to CMD is,
\begin{equation}\label{eq:CMDdyn}
    w_i^k \approx a_iw_{r,m}^k+b_i, \quad \forall i \in C_m ,
\end{equation}
with $C_m$ denoting $w_i$'s correlation cluster and $w_{r,m}$ is a reference dynamic representing all the weights in the cluster. 
A main difference between DMD and CMD modeling is exponential dynamic assumption versus data-driven dynamic, respectively. This significant difference is illustrated in Fig. \ref{cmd_vs_dmd_rec}.
We show the applicability of this approach to several popular architectures in computer vision.

\section{DMD Analysis for Neural Networks}
\label{sec:DMD}
\subsection{\acl{DMD}}
\ac{DMD} is a common modern robust tool for system analysis. It is widely applied in dynamical systems in applications ranging from fluid mechanics through finance to control systems \cite{kutz2016dynamic}. \ac{DMD} and its extensions (e.g. Exact \ac{DMD} \cite{tu2013dynamic}, tls\ac{DMD} \cite{hemati2017biasing}, fb\ac{DMD} \cite{dawson2016characterizing},
optimized \ac{DMD}  \cite{askham2018variable}, and extended \ac{DMD} \cite{williams2015data, williams2015data2}) reveal the main modes of the dynamics in various nonlinear cases and yield constructive results in various applications \cite{dietrich2020koopman}. Thus, it is natural to apply DMD-type algorithms to model the process of neural net training in order to analyze the emerging modes and to reduce its dimensions.

For instance, in \cite{naiman2021koopman} it was proposed to use \ac{DMD} as a method to analyze already trained sequential networks such as Recurrent Neural Networks (RNNs) to asses the quality of their trained hypothesis operator in terms of stability.
In \cite{manojlovic2020applications} the DMD algorithm was applied to study the stability of the training and to offer a pruning method based the operator's analysis, formulated in \cite{redman2021operator}.
The authors of \cite{tano2020accelerating} and \cite{dogra2020optimizing} propose to apply \ac{DMD} on gradient descent \ac{NN} training, and to use it for modeling or for extrapolation of further epochs. These approaches are novel and interesting. However, the examples shown are for rather unique architectures and tasks.

\ac{DMD} is originally formulated as an approximation of Koopman decomposition \cite{mezic2005spectral}, where it is argued that any Hamiltonian system has measurements which behave linearly under the governing equations \cite{koopman1931hamiltonian}. These measurements, referred to as Koopman eigenfunctions, evolve exponentially in time. This fact theoretically justifies the use of \ac{DMD}, where it can be considered as an exponential data fitting algorithm \cite{askham2018variable}.

Lately, \ac{DMD} was applied on homogeneous flows \cite{cohen2021modes}. This work bridges between \ac{DMD} and the $p$-Laplacian spectral decomposition applied in signal processing. Naive \ac{DMD} implementation on homogeneous flows was shown to be inadequate. The typical decay profile of a homogeneous flow of degree $\gamma$ is hyperbolic for $\gamma>1$ or polynomial for $0\le \gamma <1$. In the latter case one obtains a steady-state in finite time.
This contradicts the essence of \ac{DMD}, which assumes exponential decay modes. Thus, the ability of \ac{DMD} in either reconstructing the dynamic or in revealing spatial modes is limited.

These findings were recently studied via Koopman operator analysis \cite{cohen2021latent}. 
In order for Koopman mode decomposition and \ac{DMD} to coincide, the Koopman eigenfunctions must be a linear combination of the system coordinates. However, this may not happen for many nonlinear processes. In addition, the approximation by \ac{DMD} of Koopman eigenfunctions can generally not be obtained for systems with one or more of the following characteristics:
    \begin{enumerate}
        \item Finite time support.
        \item Nonsmooth dynamic - not differential with respect to time. 
        \item Non-exponential decay profiles.
    \end{enumerate}
    We give below a simple example of applying \ac{DMD} on a single-layer linear network, showing that already in this very simple case we reach the limits stated above when using augmentation.

\subsection{DMD in Neural Networks}
The training process of neural networks is highly involved. Some common practices, such as augmentation, induce exceptional nonlinear behaviors, which are very hard to model. Here we examine the capacity of \ac{DMD} to characterize a gradient descent process with augmentation. We illustrate this through a simple Toy example of linear regression.

\begin{figure}[H]
  \centering
  \begin{subfigure}{\textwidth}
  \centering
  \includegraphics[width=\linewidth]{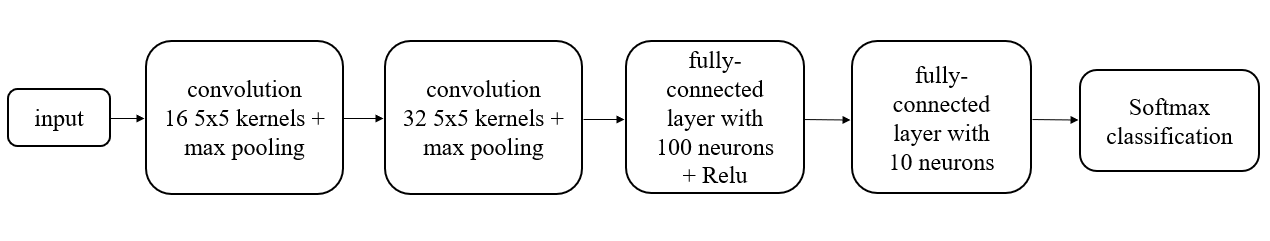}
    \caption{}
  \label{cifar10_modes_dist:sub1}
  \end{subfigure}
  \begin{subfigure}{\textwidth}
  \centering
  \includegraphics[width=\linewidth,height=40mm]{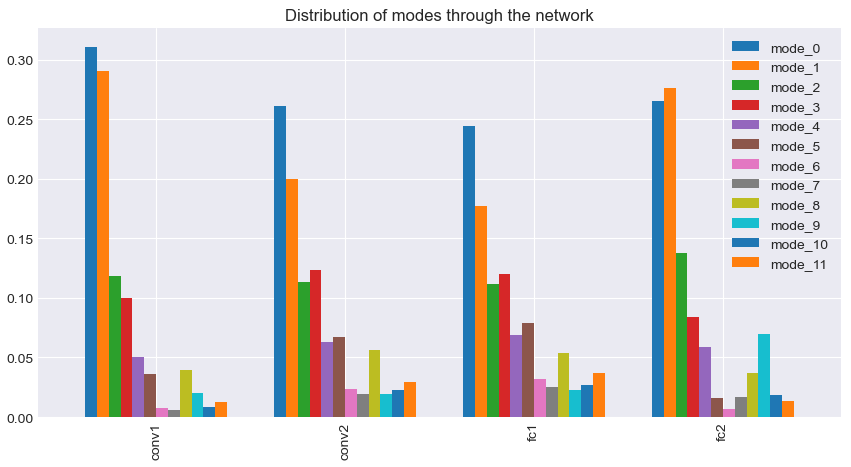}
  \caption{}
  \label{cifar10_modes_dist:sub2}
  \end{subfigure}
  \label{cifar10_modes_dist}
  \caption{(a) SimpleNet, CIFAR10 Classification. (b) Modes distribution through the Network.}
\end{figure}

\paragraph{Toy example -- linear regression with augmentation.} 
Let us formulate a linear regression problem as follows,
\begin{equation}\label{eq:regPro}
  \mathcal{L}(x,w;y) =\frac{1}{2}\norm{y-wx}_{F}^2  , 
\end{equation}
where $\norm{\cdot}_F$ denotes the Frobenius norm, $x\in \mathbb{R}^{m \times n}$, $y\in \mathbb{R}^{d \times n}$ and $w\in \mathbb{R}^{d \times m}$.
In order to optimize for $w$, the gradient descent process, initialized with some random weights $w_0$, is
\begin{equation}\label{eq:GD}
    w^{k+1}=w^k+\eta^k\left(y-w^k x\right)x^T,
\end{equation}
where the superscript $k$ denotes epoch $k$, $T$ denotes transpose and $\eta^k$ is a (possibly adaptive) learning rate.
We note that introducing an adaptive learning rate by itself already makes the dynamic nonlinear, as $\eta$ denotes a time interval.
Let us further assume an augmentation process is performed, such that at each epoch we use different sets of $x$ and corresponding $y$ matrices, that is, 
\begin{equation}\label{eq:GD_aug}
    w^{k+1}=w^k+\eta^k(y^k-w^k x^k)({x^k})^T.
\end{equation}
Then, the smoothness in Eq. \eqref{eq:GD} becomes non-smooth in the augmented process, Eq. \eqref{eq:GD_aug}. 
Actually, we obtain a highly nonlinear dynamic, where at each epoch the operator driving the discrete flow is different. We generally cannot obtain Koopman eigenfunctions and cannot expect \ac{DMD} to perform well. 
Our experiments show that this is indeed what happens in practice. In the modeling of simple networks, \ac{DMD} can model the dynamics of classical gradient descent but fails as soon as augmentation is introduced. Figure \ref{cifar10_gd_cmd_res:sub2} presents an experiment of CIFAR10 classification using a simple CNN (Fig \ref{cifar10_modes_dist:sub1}) with standard augmentations, usually performed on this dataset (horizontal flips and random cropping). In this experiment in order to reconstruct the dynamics, we perform DMD with different dimensionality reduction rates ($r = 10, 50, 90$) using Koopman node operator, as described in \cite{dogra2020optimizing}.
The results presented in Figure \ref{cifar10_gd_cmd_res:sub3} show that when the dimensionality reduction is  high ($r = 10$) DMD fails to reconstruct the network's dynamics. However, when the dimensionality reduction is mild  ($r = 90$) DMD reconstruction is also unstable and oscillatory. Similar arguments hold for stochastic gradient descent (SGD). Therefore, our aim is to seek an alternative low dimensional representation of the dynamics, which can handle strong nonlinearities over a large and complex network, as well as standard training practices, like augmentation and SGD.
Another failure case we have encountered was using the DMD modeling for complex architectures even without augmentations, such as Resnet18 (Fig. \ref{cifar10_gd_cmd_res:sub3}). More details of this experiment are given in the supplementary material.

\begin{figure}
  \centering
  \begin{subfigure}{0.33\textwidth}
  \centering
  \includegraphics[height=40mm]{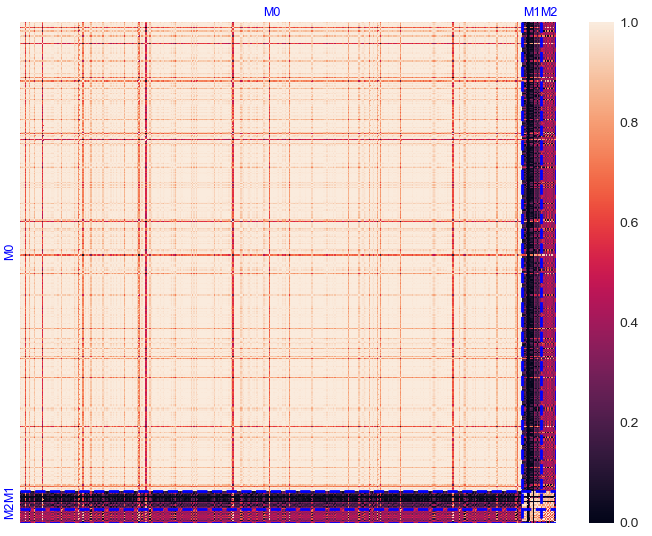}
  \caption{SimpleNet, Binary classification}
  \label{clustered_corr_matrices:sub1}
  \end{subfigure}
  \begin{subfigure}{0.33\textwidth}
  \centering
  \includegraphics[height=40mm]{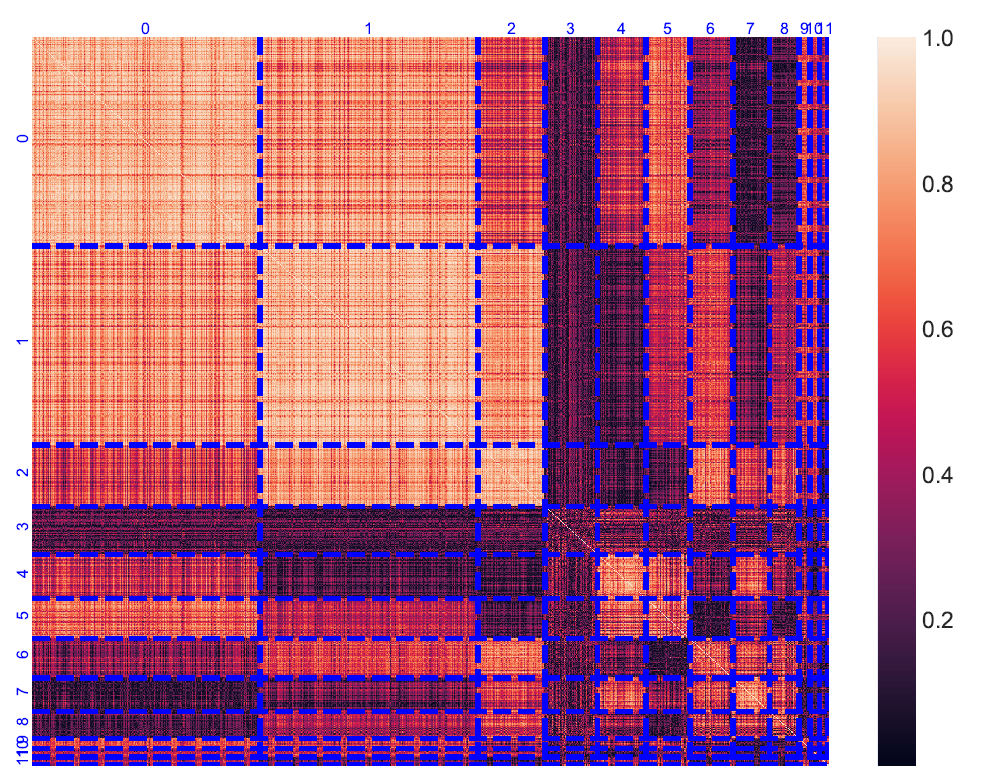}
  \caption{SimpleNet, CIFAR10 classification}
  \label{clustered_corr_matrices:sub2}
  \end{subfigure}
  \begin{subfigure}{0.32\textwidth}
  \centering
  \includegraphics[height=40mm]{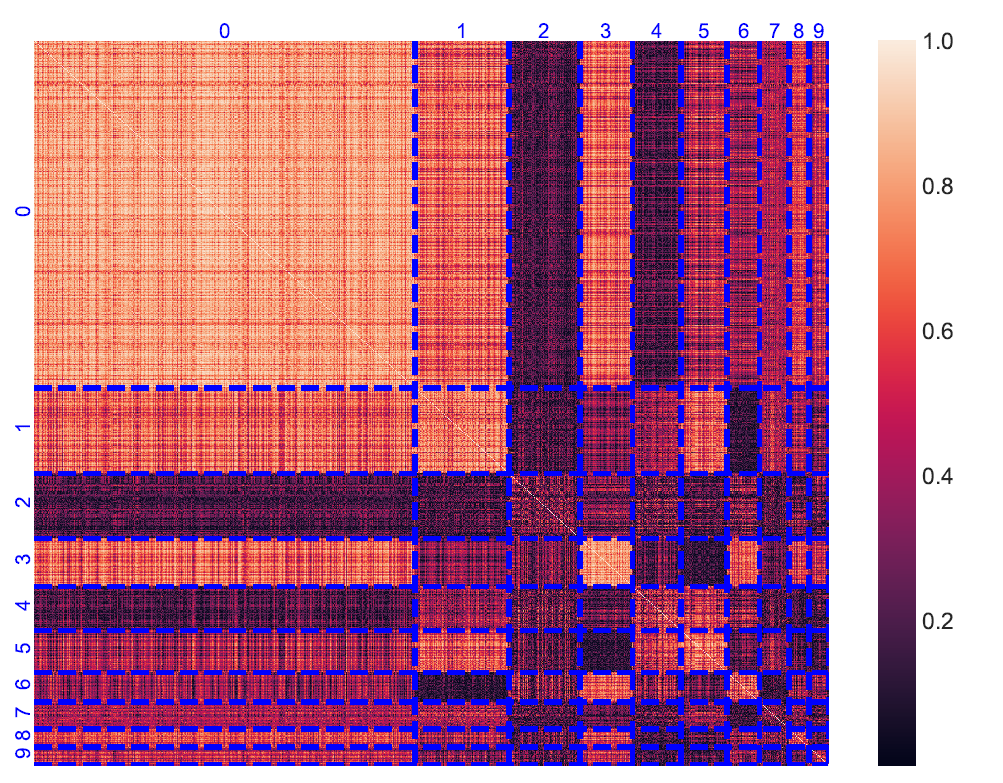}
  \caption{Resnet18, CIFAR10 classification}
  \label{clustered_corr_matrices:sub3}
  \end{subfigure}
  \caption{The clustered correlation matrix of the sampled weights. (a) SimpleNet, Binary classification. (b) SimpleNet, CIFAR10 classification.
  (c) Resnet18, CIFAR10 classification.
Dashed blue lines separate the modes, the mode numbers are shown on the top and left margins of the matrix.}
  \label{clustered_corr_matrices}
\end{figure}

\begin{figure}
  \centering
    \begin{subfigure}{.3\textwidth}
  \centering
  \includegraphics[width=\linewidth,height=45mm]{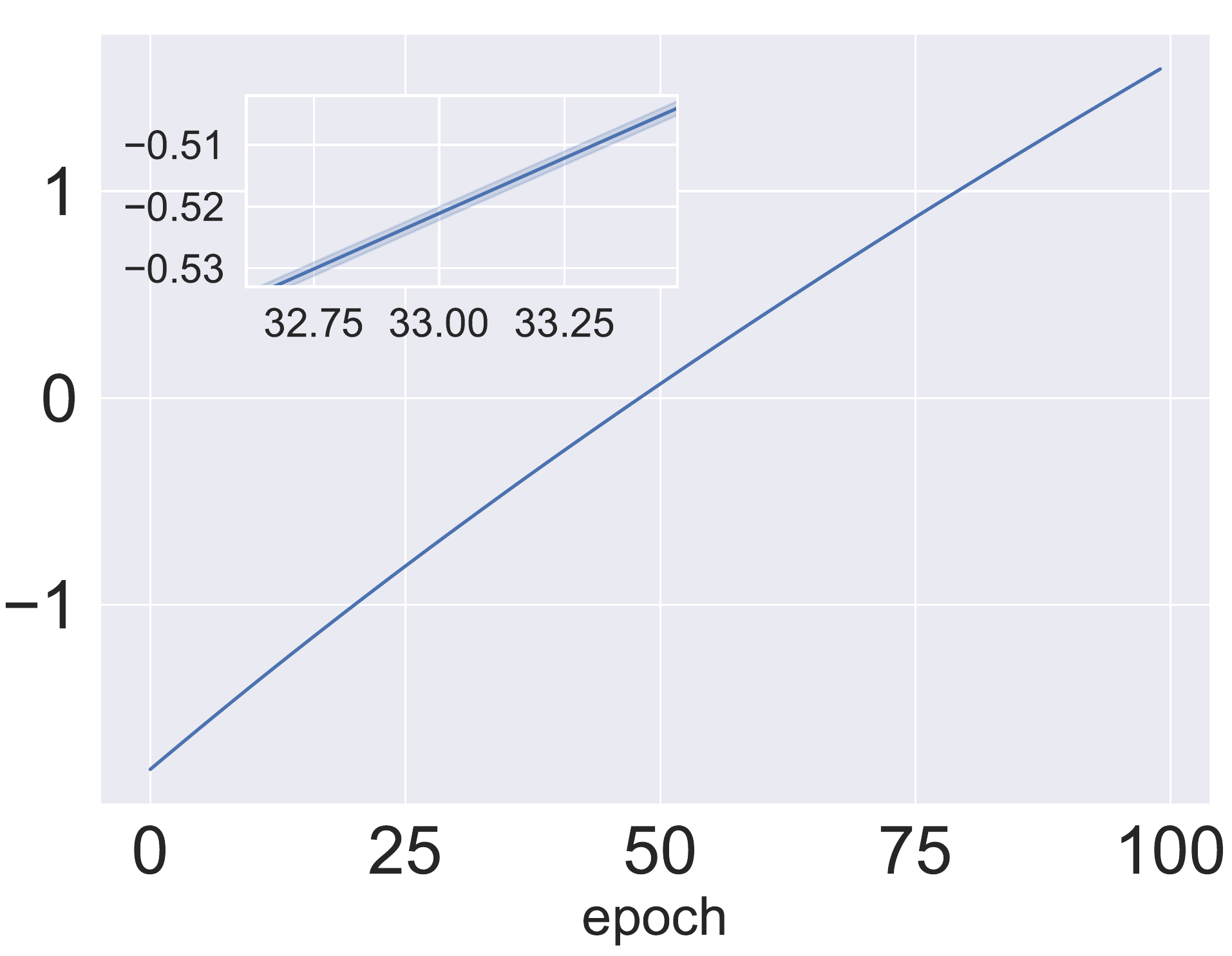}
  \caption{Mode 0}
  \label{toy_modes_95conf_interval:sub1}
  \end{subfigure}
\begin{subfigure}{.3\textwidth}
  \centering
  \includegraphics[width=\linewidth,height=45mm]{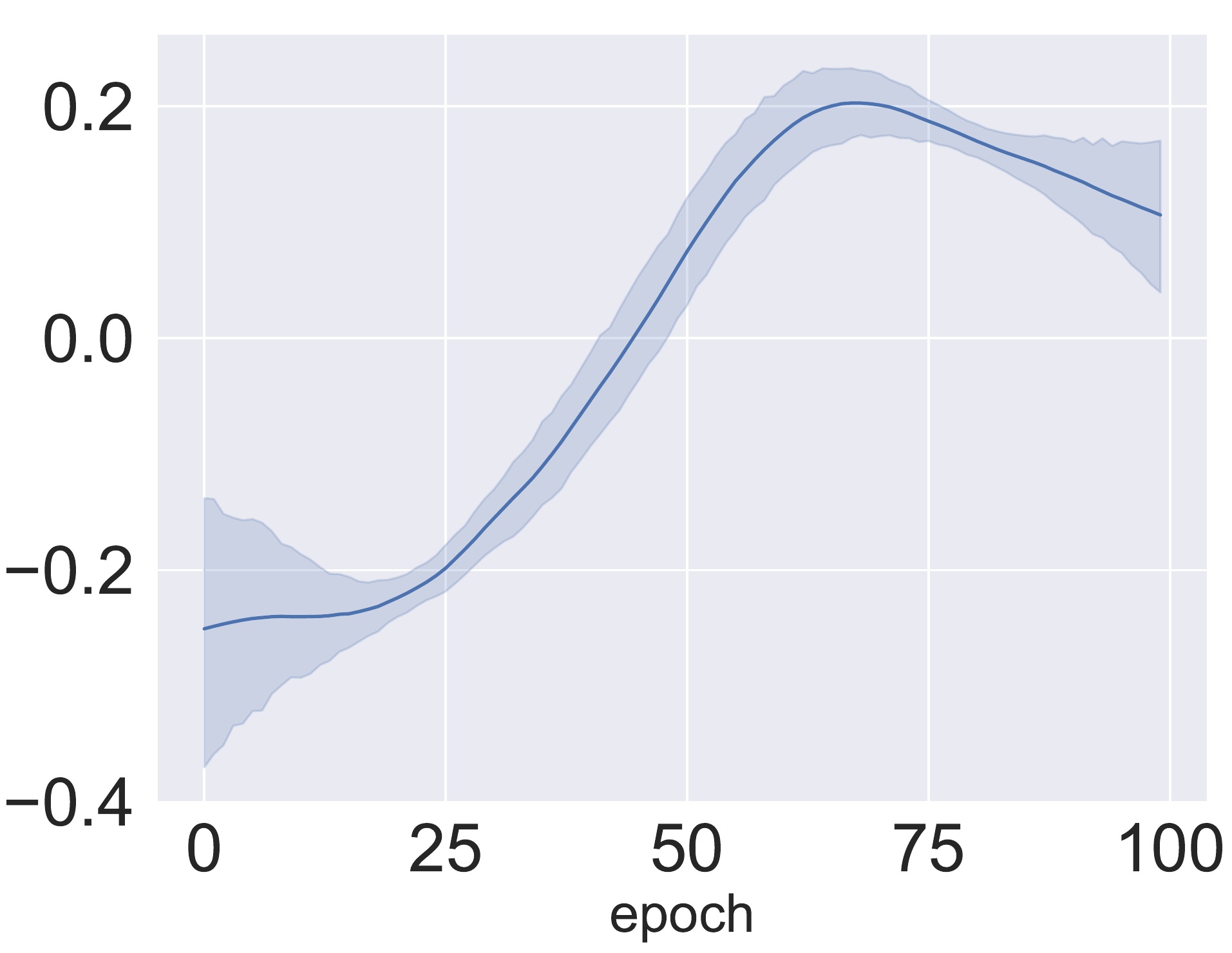}
  \caption{Mode 1}
  \label{toy_modes_95conf_interval:sub2}
  \end{subfigure}
  \begin{subfigure}{.3\textwidth}
  \centering
  \includegraphics[width=\linewidth,height=45mm]{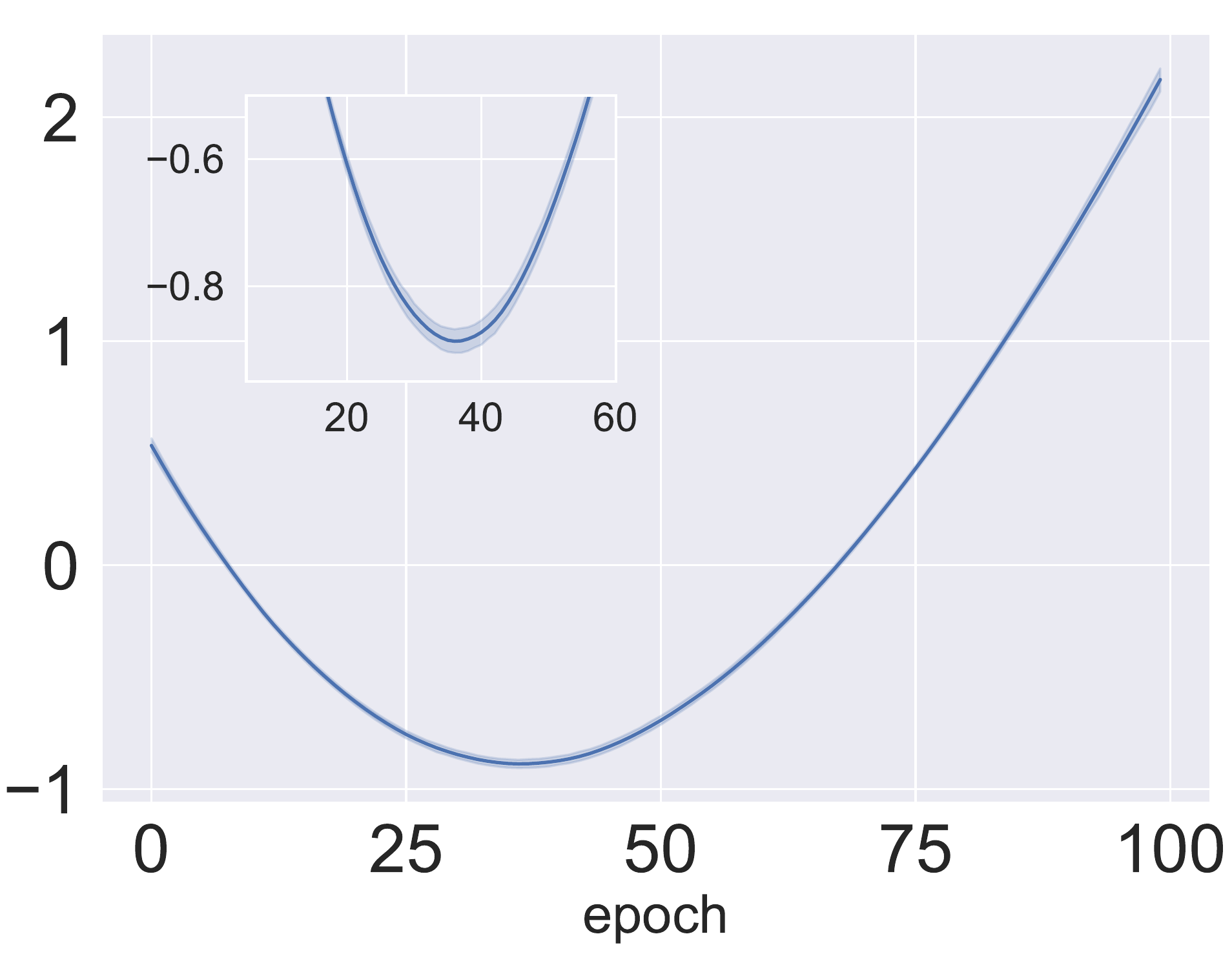}
  \caption{Mode 2}
  \label{toy_modes_95conf_interval:sub3}
  \end{subfigure}
  \caption{95\% confidence interval of the modes. For Mode 0 and Mode 2 the interval is very dense (highly concentrated cluster), thus we show an enlargement at the top of the plot. }
  \label{toy_modes_95conf_interval}
\end{figure}

\section{Proposed method}
Our approach is based on the analysis of time-profiles, which in the neural-network setting is equivalent to examining the behavior of the network parameters, as they evolve through epochs of gradient descent. Aubry et al \cite{aubry1991spatiotemporal,aubry1992spatio} have shown (already in the 90's) that time-profiles and correlation analysis is beneficial in modeling nonlinear physical 
phenomena. Their aim was to decompose the dynamic to orthogonal components both in space and in time.
Imposing orthogonality in space and time, however, may be too strong of a constraint, leading to a limited solution space.

More recently, studies such as \cite{burger2016spectral,gilboa2018nonlinear} in variational image-processing have shown that gradient descent with respect to homogeneous functionals (of various degrees) induce typical time profiles. For instance, total-variation flow \cite{andreu2001minimizing} can be modelled by piecewise linear time profiles. These profiles stem from the behavior of basic elements with respect to the gradient operator, referred to as \emph{nonlinear eigenfunctions}.
The theory developed there shows that the time profiles are generally not orthogonal. Orthogonality was shown in certain settings for the spatial structures (``spectral components''). We attempt to generalize these concepts for the neural network case.

A principal difference in the modeling is that unlike the variational case, here we have no guaranteed homogeneity and the system is too complex to be modeled analytically. Neither can we count on pre-defined time profiles (e.g. exponents) or orthogonality in space as assumed for DMD. Thus we resort to data-driven time-profiles, which change with network architecture and learning task. With respect to orthogonality of the spatial structures, it is unclear how to measure it, since there is no natural inner-product over the network parameters space. Our simplistic ``zero-order'' solution is to have disjoint sets of parameters. Thus we would like to group (cluster) parameters, such that each group has a different prototypical time-profile. We refer to each such group as \emph{Mode}.

\subsection{Analysis of a simple Convolutional Neural Network (CNN)}
Let us consider a simple problem of binary image classification between cats and dogs, using a "toy" CNN. The network has several convolution layers, followed by max-pooling, fully-connected (FC) and Relu activation. It has a total of 94,000 parameters. We refer to it as \emph{SimpleNet2}, The architecture is as in Fig. \ref{cifar10_modes_dist:sub1}, except for the last layer which has one neuron instead of ten.

Training is performed in a straightforward manner, by simple gradient-descent,  without SGD, augmentation or any other advanced training scheme. Learning rate is 0.001 without scheduling. 

When examining the weights evolution during the training process for that problem, we have noticed that the general characteristics of the profiles are very similar throughout the net. Following normalization of the mean, variance and sign – there are essentially very few characteristic profiles which represent the entire dynamics. Moreover, these profiles are spread throughout the network and can be extracted by uniform sampling of a small subset of the entire network parameters. To illustrate this, we sampled 1000 weights of the network parameters ($\approx 1\%$) and clustered them into 3 main modes, based on the correlations between the weights in this subset. In Figure \ref{clustered_corr_matrices:sub1}, one can easily notice that the weights are indeed divided into highly distinct modes. We then took the rest of the CNN's parameters and associated each of them to the mode it is most correlated to. 

A good way to visualize the correlation within the modes is to plot the $95\%$ confidence interval. For each mode we chose a reference parameter, which is most correlated to all other parameters in the cluster (``cluster center''). At each epoch we compute the range of the $95\%$ confidence interval, a plot is shown in Fig. \ref{toy_modes_95conf_interval}. It is clear that Mode 0 and Mode 2 are highly condensed clusters, whereas Mode 1 is more spread-out.
We can now state our \emph{Correlation Mode Decomposition} (CMD) algorithm.

\subsection{CMD Algorithm}\label{sec:cmd_alg}
 {\bf Model hypothesis:} \emph{The dynamics of parameters of a neural network can be clustered into very few highly correlated groups (Modes).}

 Let us formalize this. Let $N$ be the number of network parameters,  $M \ll N$ the number of clusters $\{C_1,..\,\,,C_M\}$ and $0 <\epsilon \ll 1$ a small threshold parameter. Then 
\begin{equation}
\label{eq:model_corr}
    |\textrm{corr}(w_i,w_j)| \ge 1-\epsilon, \quad \forall w_i,w_j\in C_m, \quad m=1..\, M.
 \end{equation}
 The correlation is defined as $ \textrm{corr}(u,v) = \frac{\langle \Bar{u},\Bar{v} \rangle}{\norm{\Bar{u}}\norm{\Bar{v}}}$, with $\Bar{u}$ denoting a centralized weight by $\Bar{u} = u-\frac{1}{T+1}\sum_{k=0}^{T}{u^k}$, and $\langle \cdot , \cdot \rangle$ is the Euclidean inner product over the epoch axis $\langle u, v\rangle = \sum_{k=0}^{T}{u^k v^k}$.
 Thus, any two signals $s_1^k,s_2^k$ which are perfectly correlated (or anti-correlated), yielding  $|\textrm{corr}(s_1,s_2)| = 1$, can be expressed as an affine transformation of each other: $s_2^k=a\cdot s_1^k+b $, where $a,b\in \mathbb{R}$. 
This leads to the following approximation of the dynamics:    
\begin{equation}
\label{eq:model_weights}
    w_i^k \approx a_i\cdot w_r^k + b_i, \quad \forall w_i, w_r\in C_m, \quad m=1..\, M,
\end{equation} 
where $w_r$ is a reference weight in the cluster and $a_i$, $b_i$ are affine coefficients corresponding to each  weight.
The clustering of the parameters into modes is given in Algorithm \ref{alg:linear_corr_clust}. 
Weight reconstruction by CMD is given in Algorithm \ref{alg:cmd}.

There are two options for choosing the number of modes $M$:
\begin{enumerate}
    \item Predetermined M: Finding a minimum threshold $t$ so that the cophenetic distance between any two original observations in the same cluster is no more than $t$, where no more than $M$ clusters are formed.
    \item In-cluster distance threshold: Forming clusters so that the original observations in each cluster have no greater a cophenetic distance than a desired threshold $t$.
\end{enumerate}

In order to find $a$ and $b$ we propose the following computation,
\begin{equation}
\label{eq:AB}
    \{A, B\} = \argmin_{A,B}{\|W_m - Aw_{r,m} + B\mathbbm{1}\|_\mathcal{F}^2},
\end{equation}
where $W_m \in \mathbb{R}^{|C_m|\times T}$ is a matrix of all weight dynamics in the mode $C_m$, $A \in \mathbb{R}^{|C_m|\times 1}$ is the vector of coefficients $a_i$, $B \in  \mathbb{R}^{|C_m|\times1}$ is the vector of free terms $b_i$ and $\mathbbm{1}\in \mathbb{R}^{1\times T}$ is a row vector of ones. By defining the matrix $\tilde{A}:=[A\: B]$ and $\tilde{w}_{r,m}:=\begin{bmatrix} w_{r,m}  \\ \mathbbm{1}  \end{bmatrix}$ 
we get the relation,
\begin{equation}
\label{eq:A_tilde_min}
    \tilde{A} = \argmin_{\tilde{A}}{\|W_m-\tilde{A}\tilde{w}_{r,m}\|^2_\mathcal{F}},
\end{equation}
with the solution
\begin{equation}
    \label{eq:A_tilde}
    \tilde{A} = W_m\tilde{w}_{r,m}^T\left(\tilde{w}_{r,m}\tilde{w}_{r,m}^T \right)^{-1}.
\end{equation}

\begin{algorithm}
\caption{Correlation-based clustering with linear complexity }\label{alg:linear_corr_clust}
\begin{algorithmic}
\State \textbf{Input:}
\begin{itemize}
    \item[-] $W\in \mathbb{R}^{N\times T}$ - Matrix of all weights.
    \item[-] $K$ - Number of representative weights.
    \item[-] $M$ / $t$ - Number of wanted modes / desired in-cluster distance threshold.
\end{itemize}
\State \textbf{Output:}
\begin{itemize}
    \item[-] $\{C_m\}_{m=1}^{M}$ - Clusters of the network parameters.
\end{itemize}

\Procedure {}{}
\begin{enumerate}
    \item Initialize $C_m= \{\}$,  $m=1..M$.
    \item Sample a subset of $K$ random weights and compute their correlations.
    \item Cluster this set to $M$ modes based on correlation values, update $\{C_m\}$ accordingly.
    \item Choose profiles $w_{r,m} = \argmax_{w_j \in C_m} \sum_{w_i\in C_m}|\textrm{corr}(w_i, w_j) |$.
\end{enumerate}

\For{$i$ in ($N-K$) weights}
    \State $m^* = \argmax_m |\textrm{corr}(w_i, w_{r,m}) |$
    \State $C_{m^*} \leftarrow C_{m^*} \cup \{w_i\}$
\EndFor
\State \textbf{return $\{C_m\}_{m=1}^{M}$}
\EndProcedure
\end{algorithmic}
\end{algorithm}

\subsection{Correlation-based clustering with linear complexity} Estimating the correlation between 
$N$ variables typically needs an order of $N^2$ computations (every variable with every other variable). This can be problematic for large networks where $N$ can be in the order of millions (or even billions).
However, with our model (summarized in Eqs. \eqref{eq:model_corr}, \eqref{eq:model_weights})
 clustering can be performed without computing the full correlation matrix and is essentially in the order of $N\cdot M \cdot T$, where $M$ is the number of modes and $T$ is the number of epochs. Instead of computing the entire correlation matrix, we propose to compute the correlations only between the network weights and the reference weights of each modes, which were found earlier in the sampling phase.
The estimation procedure is described in Algorithm \ref{alg:linear_corr_clust}.
The complexity is $K^2\cdot T +(N-K)\cdot M \cdot T \approx N\cdot M \cdot T$. $K$ can be in the order of $30 \times M$ to provide sufficient statistics. In all our experiments we use $K=1000$. An implicit assumption is that in the $K$ sampled weights we obtain all the essential modes of the network.

\begin{algorithm}
\caption{CMD - Correlation Mode Decomposition}\label{alg:cmd}
\begin{algorithmic}
\State \textbf{Input:}
\begin{itemize}
    \item[-] $W\in \mathbb{R}^{N\times T}$ - Matrix of all weights.
    \item[-] $K$ - Number of sampled weights.
    \item[-] $M$ / $t$ - Number of wanted modes / desired in-cluster distance threshold.
\end{itemize}
\State \textbf{Output:}
\begin{itemize}
    \item[-] $\hat{W} \in \mathbb{R}^{N\times T}$ - CMD reconstruction of network parameters.
\end {itemize}

\Procedure {CMD}{}
\State Perform correlation clustering to $M$ modes using Alg. \ref{alg:linear_corr_clust}. ($M$ is fixed if given, or deduced according to the $t$ input). Get mode mapping $\{C_m\}_{m=1}^{M}$. 
\For{$m \gets 1$ to $M$}
    \State Follow Eqs. \eqref{eq:AB}, \eqref{eq:A_tilde_min} and \eqref{eq:A_tilde} to compute $A, B$.
    \State $\hat{W}_m \gets Aw_{r,m}+B\mathbbm{1}$, where $\hat{W}_m$ is the CMD reconstruction of $W_m$.
\EndFor
\State \textbf{return $\hat{W} = \cup_{m=1}^{M}\{\hat{W}_m$\}}
\EndProcedure
\end{algorithmic}
\end{algorithm}

\begin{figure}
  \centering
  \begin{subfigure}{.28\textwidth}
  \centering
  \includegraphics[width=\linewidth,height=40mm]{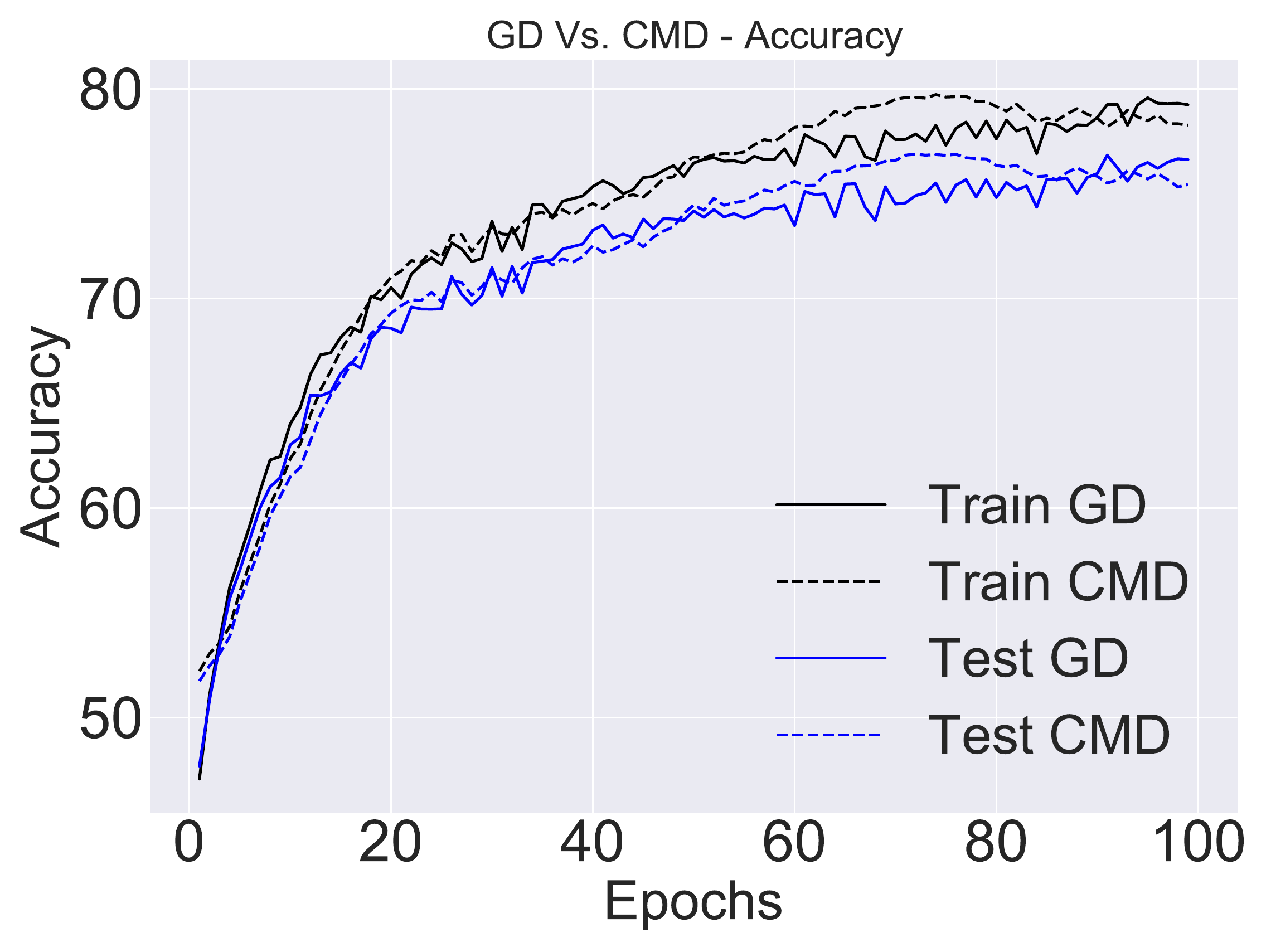}
  \caption{CMD modeling}
  \label{cifar10_gd_cmd_res:sub1}
  \end{subfigure}
     \begin{subfigure}{.32\textwidth}
  \centering
     \includegraphics[width=\linewidth,,height=40mm]{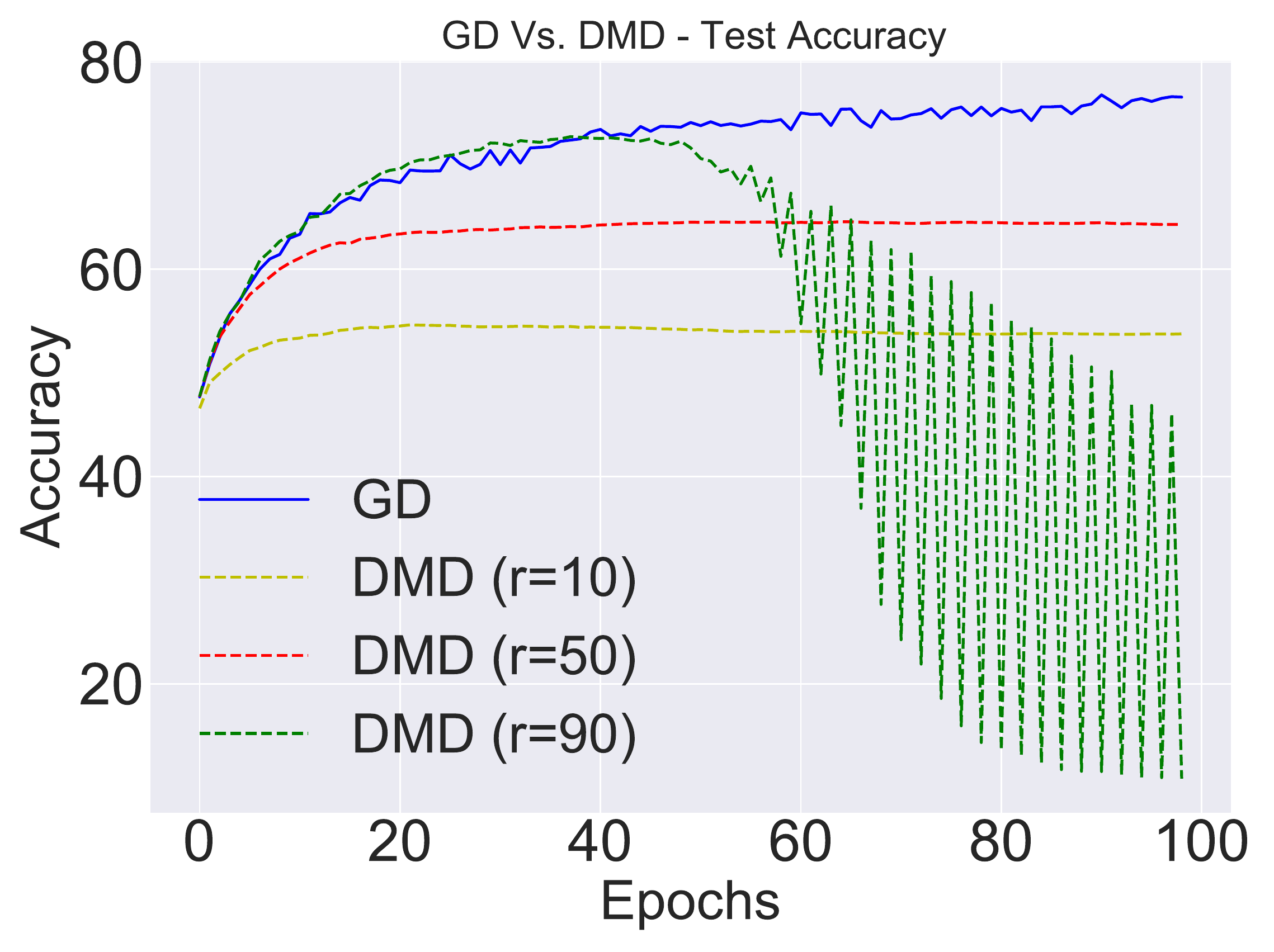}
  \caption{DMD failure with augmentations}
  \label{cifar10_gd_cmd_res:sub2}
  \end{subfigure} 
  \begin{subfigure}{.28\textwidth}
  \centering
  \includegraphics[width=\linewidth, height=40mm]{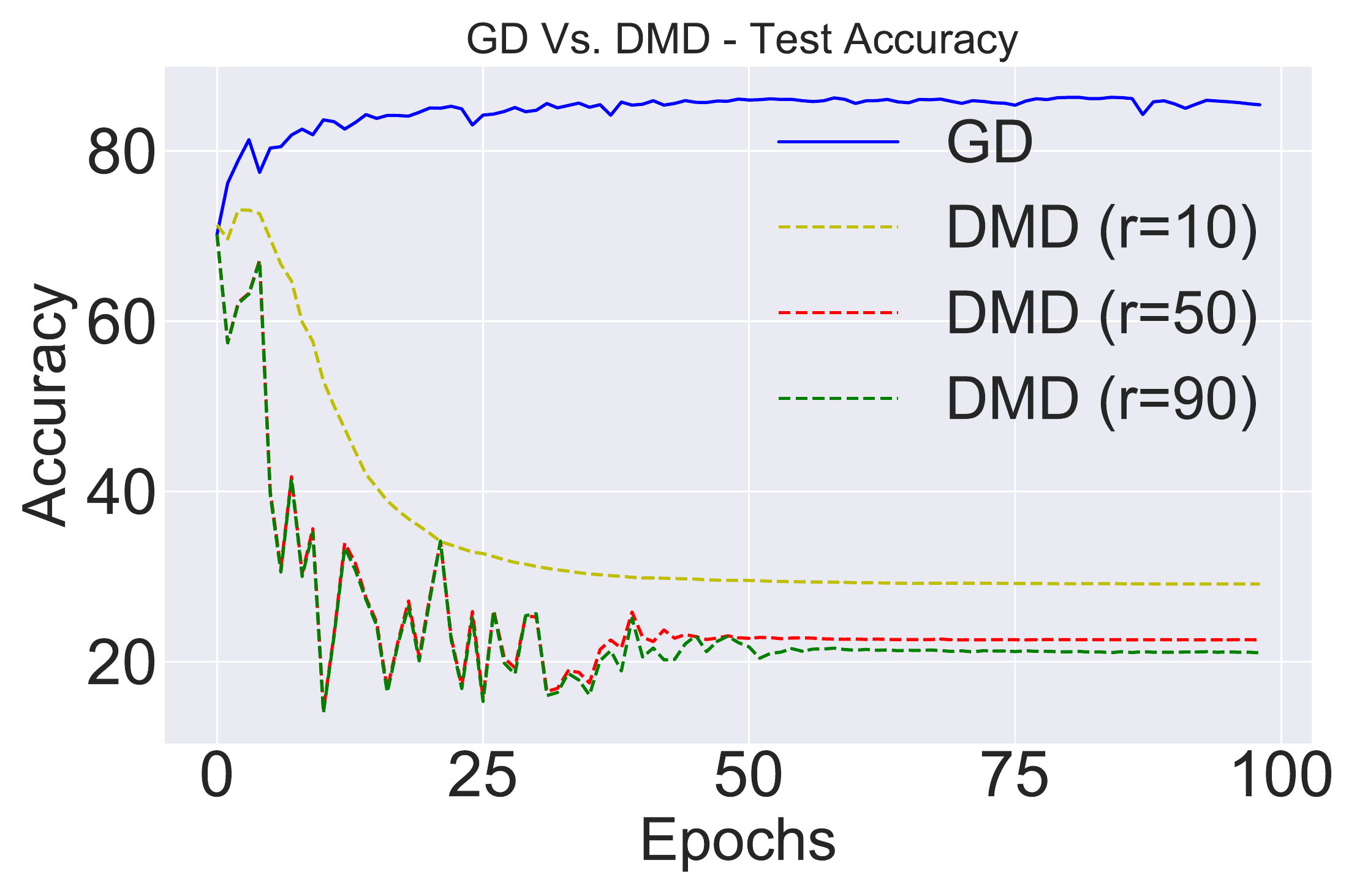}
  \caption{DMD failure with complex architectures}
  \label{cifar10_gd_cmd_res:sub3}
  \end{subfigure}
  \caption{CIFAR10 classification results, Original Gradient Descent (GD) training Vs. DMD modeling and CMD modeling. \textbf{(a)} CMD modeling, SimpleNet architecture, with augmentations. \textbf{(b)} DMD modeling, SimpleNet architecture, with augmentations (3 values of dimensionality reduction parameter $r$). \textbf{(c)} DMD modeling, Resnet18, no augmentations.}
  \label{cifar10_gd_cmd_res}
\end{figure}

\begin{figure}
  \centering
  \begin{subfigure}{.47\linewidth}
  \centering
  \includegraphics[width=\linewidth,height=40.5mm]{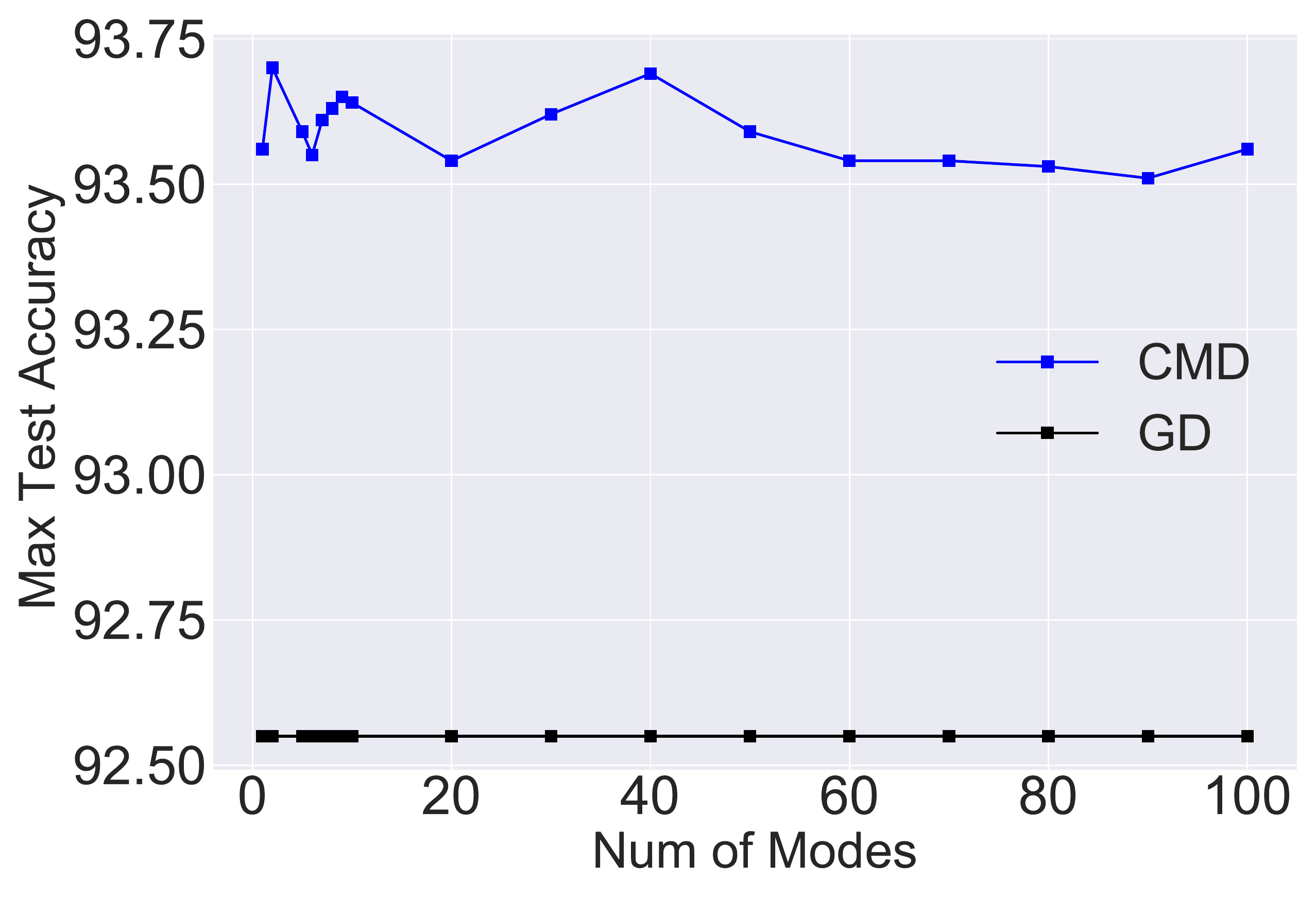}
  \caption{Accuracy}
  \label{performance_vs_modes_cifar10_resnet18:sub1}
  \end{subfigure}
  \begin{subfigure}{.52\linewidth}
  \centering
  \includegraphics[width=\linewidth,height=39mm]{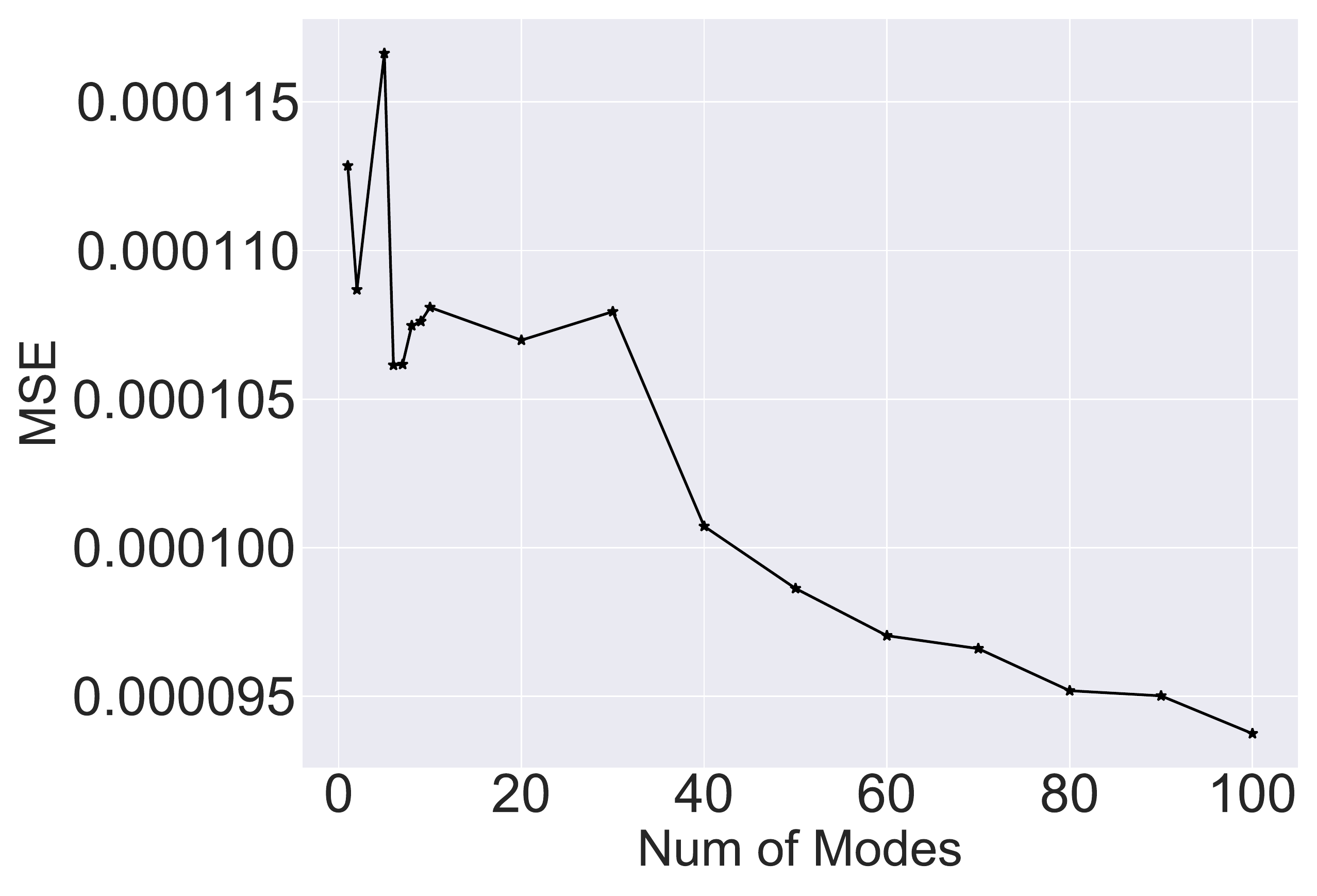}
  \caption{Weights MSE}
  \label{performance_vs_modes_cifar10_resnet18:sub2}
  \end{subfigure}
  \caption{Resnet18 CIFAR-10 results (a) Maximal test Accuracy of CMD model Vs. Number of modes used, for $M = 1,2,3..10,20,30,..100$. (b) MSE of Weights reconstruction Vs. Number of modes used for CMD.}
  \label{performance_vs_modes_cifar10_resnet18}
\end{figure}

\begin{figure*}
  \centering
  \begin{subfigure}{.49\linewidth}
  \centering
  \includegraphics[width=\linewidth,height=43mm]{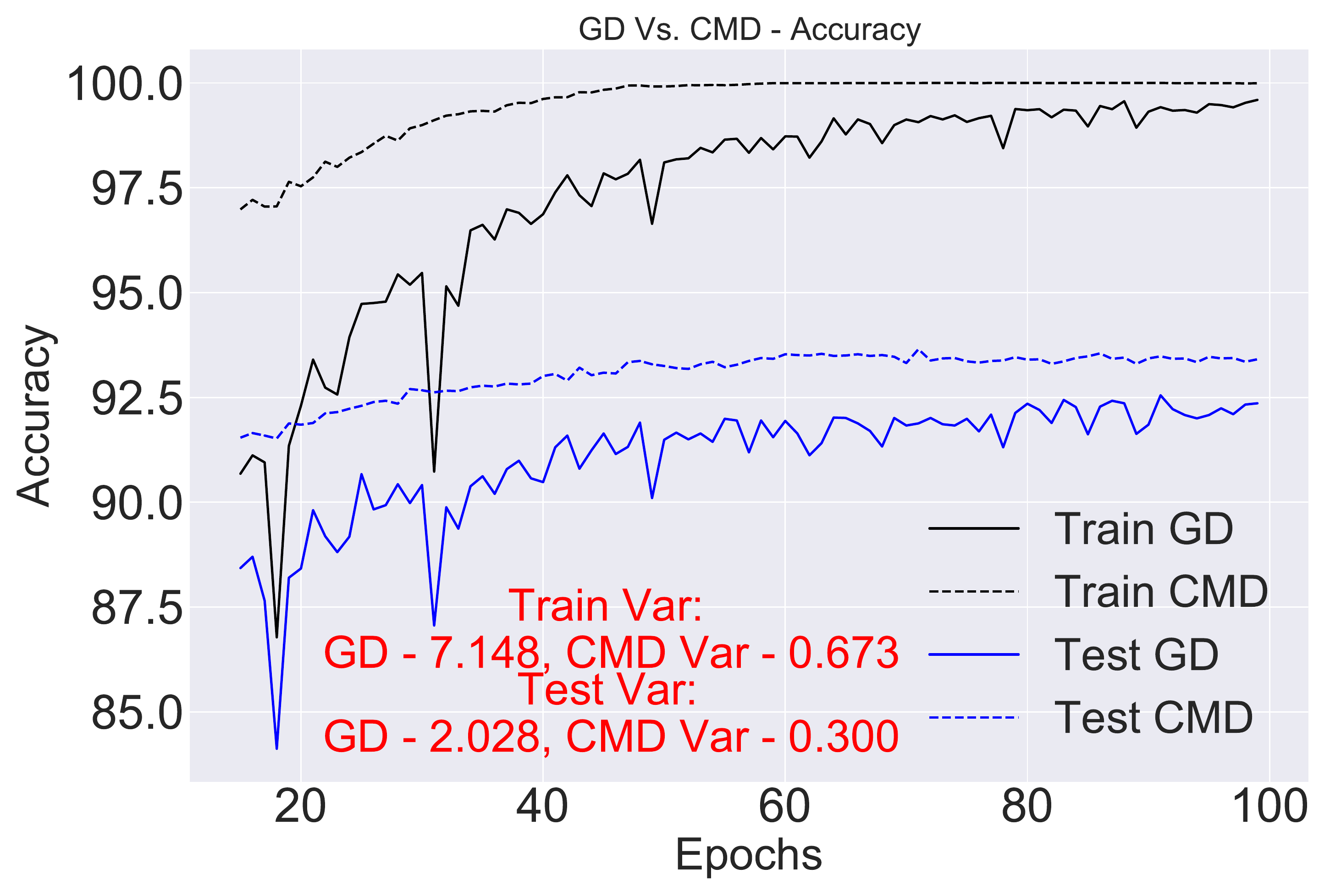}
  \caption{ResNet18}
  \label{resnet_vit_gd_cmd_res:sub1}
  \end{subfigure}
  \begin{subfigure}{.49\linewidth}
  \centering
  \includegraphics[width=\linewidth,height=47mm]{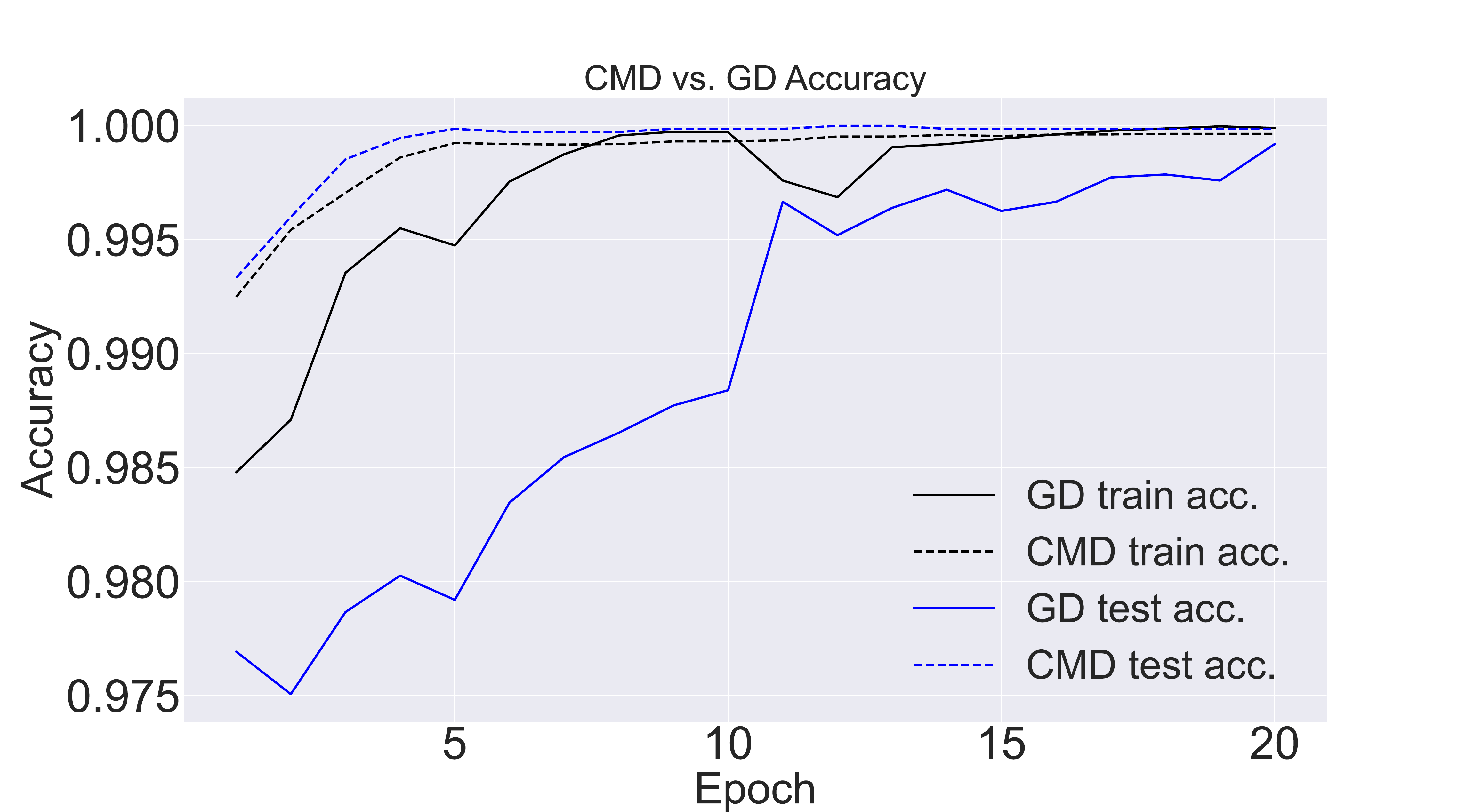}
  \caption{ViT-b-16}
  \label{resnet_vit_gd_cmd_res:sub2}
  \end{subfigure}
  \caption{CIFAR10 classification results - Accuracy of Original Gradient Descent (GD) training Vs. CMD modeling. (a) ResNet18. (b) ViT-b-16.
  }
  \label{resnet_vit_gd_cmd_res}
\end{figure*}

\section{Experiments}
\label{sec:experiments}
We present several experimental results on different networks which show the validity of the proposed model.

\paragraph{A Small Network -- SimpleNet}
First we consider the CIFAR10  \cite{krizhevsky2010cifar} classification problem, using a simple CNN architecture, used in \cite{manojlovic2020applications}, we refer to it as \emph{SimpleNet} (Fig. \ref{cifar10_modes_dist:sub1}).

It uses CE Loss and augmented data (horizontal flip, random crop, etc.).
The CNN was trained for 100 epochs, using Adam optimizer with momentum, and initial learning rate of $\eta=1\times 10^{-3}$.
The CMD Analysis was applied on the snapshots of this training.

In this case we obtained 12 modes. The rest of the weights were extrapolated from those modes as described in Algorithm \ref{alg:cmd}. The results from the CMD and the original GD training are presented in Fig. \ref{cifar10_gd_cmd_res:sub1}, and compared to the performance of DMD \ref{cifar10_gd_cmd_res:sub2}, as detailed in Section \ref{sec:DMD}. In addition, the distribution of the modes through the entire network is presented in Figure \ref{cifar10_modes_dist:sub2}.
We clearly see the model follows well the GD dynamic, where test accuracy is even slightly higher. 

\paragraph{Image Classification using ResNet 18}
In order to verify that our method can model larger architectures, the same experiment was performed on CIFAR10 classification using ResNet18 \cite{he2016deep}.
We ran several experiments, examining the influence of the number of modes $M$ in Algorithm \ref{alg:cmd}. Figure \ref{performance_vs_modes_cifar10_resnet18:sub1} presents the  maximum accuracy of the CMD model on the test set as function of number of modes used in the CMD algorithm. The dynamic of the refernce weights for two, five and ten modes is presented in the supplementary material.
Surprisingly, the test accuracy is close to invariant to the number of modes. We reach the highest score for only two modes. Even a single mode can work, but the behavior is somewhat less stable. It appears ResNet is especially well suited for our proposed model. The performance of CMD (10 modes) is presented in Figure \ref{resnet_vit_gd_cmd_res:sub1}.

\paragraph{Image Segmentation using PSPNet}
Our method generalizes well to other vision tasks. In this section we present CMD modeling on segmentation task for PASCAL VOC 2012 dataset \cite{everingham2010pascal}.

\begin{figure}
  \centering
  \begin{subfigure}{.49\linewidth}
  \centering
  \includegraphics[width=\linewidth, height=35mm]{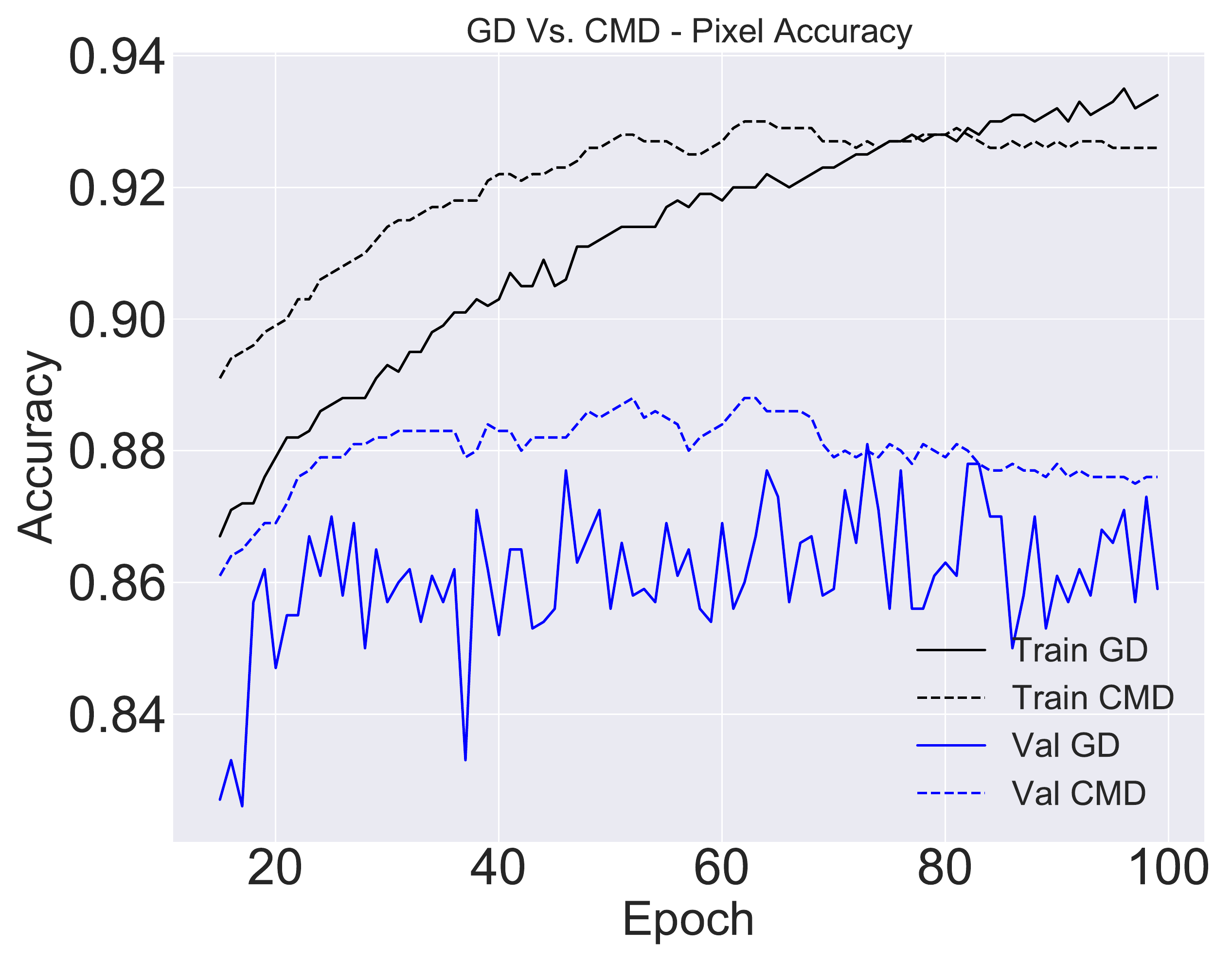}
  \caption{Pixel Accuracy}
  \label{pspnet_gd_cmd_res:sub1}
  \end{subfigure}
  \begin{subfigure}{.49\linewidth}
  \centering
  \includegraphics[width=\linewidth, height=35mm]{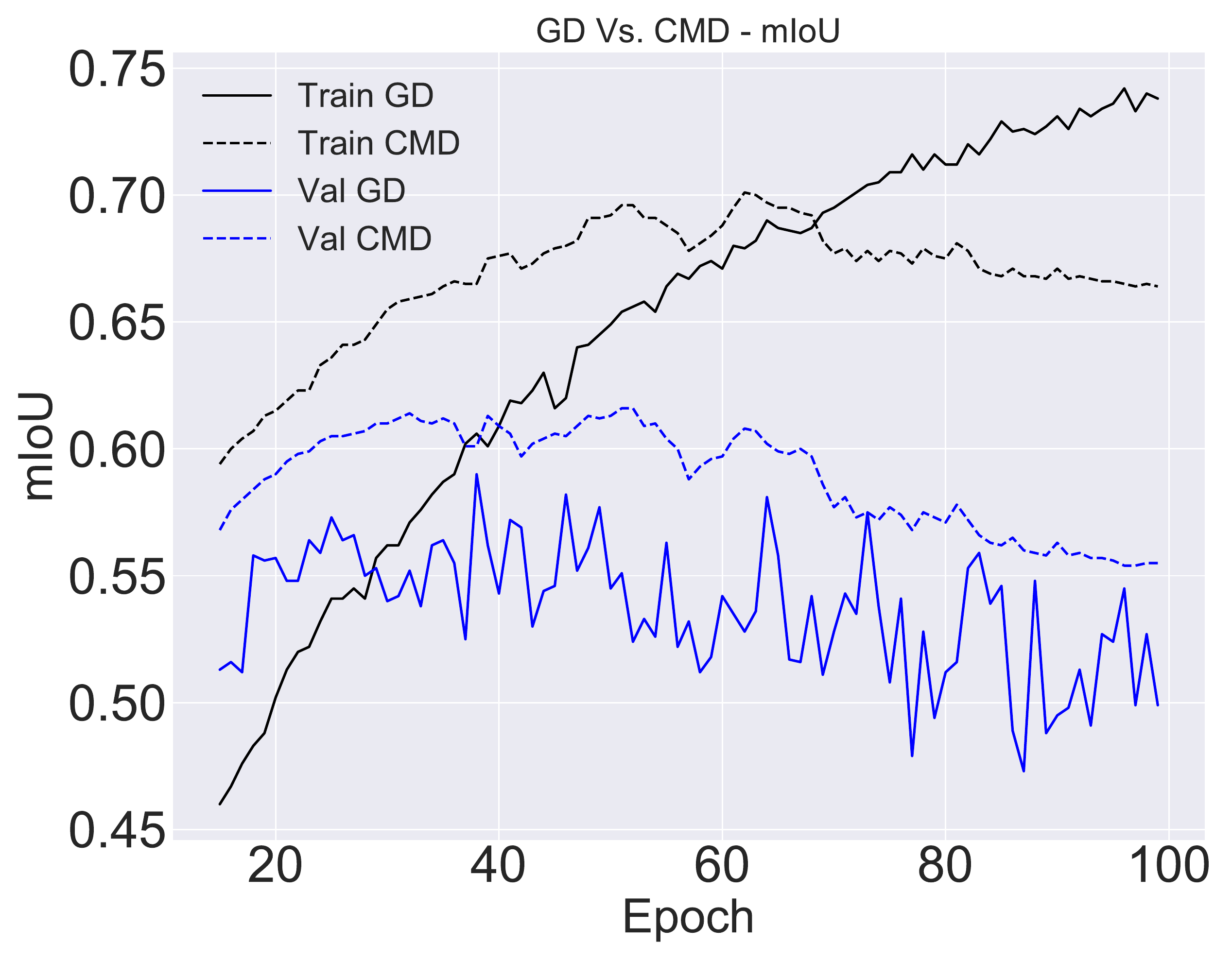}
  \caption{mIoU}
  \label{pspnet_gd_cmd_res:sub3}
  \end{subfigure}
  \caption{PASCAL VOC 2012 Segmentation results, PSPNet Architecture: Original Gradient Descent (GD) training Vs. CMD modeling. (a) pixel accuracy. (b) mIoU. \textbf{Max CMD performance on validation set:} Pixel accuracy: 88.8\% with mIoU of 61.6\%. \textbf{Max GD performance on validation set:} Pixel accuracy: 88.1\% with mIoU of 58.9\%}
  \label{pspnet_gd_cmd_res}
\end{figure}

For that task we used PSPNet Architecture \cite{zhao2017pyramid} ($13\times10^{6}$ trainable parameters). The model was trained for 100 epochs, using SGD optimizer with momentum of 0.9, and weight decay of $1\times 10^{-4}$. In addition we used “poly” learning policy with initial learning rate of $\eta=1\times10^{-2}$. The CMD Algorithm \ref{alg:cmd} was executed using $M=10$ modes. Pixel accuracy and mIoU results from the CMD modeling and the original GD training are presented in Figure \ref{pspnet_gd_cmd_res}. We observe CMD follows well GD in the training, until it reaches the overfitting regime. For the validation set, CMD is more stable and surpasses GD for both quality criteria. 

\paragraph{Classification using a Vision Transformer}
In recent years, after becoming a key tool in NLP \cite{devlin2018bert} transformers are making a large impact in the field of computer vision. Following  \cite{dosovitskiy2020image}, we apply our method on a pre-trained vision transformer. 
Our model was applied on the fine tuning process of a pre-trained ViT-b-16 \cite{dosovitskiy2020image, git_HuggingFace_Accelerate}on CIFAR10 with 15\% validation/training split.
We used Adam optimizer with a starting learning rate of $5\times 10^{-5}$ with linear scheduling. The network contains 86 million parameters. Negative Log Likelihood Loss was used. One can see in Fig. \ref{resnet_vit_gd_cmd_res:sub2} that our method models the dynamics  well and the induced regularization of CMD yields a stable, non-oscillatory evolution. 

\paragraph{Image synthesis using StarGan-v2}
We applied our CMD modeling on image synthesis task using StarGan-V2 \cite{choi2020stargan}. This framework consists of four modules, each one contains millions of parameters. It was modeled successfully by only a few modes. Details of the experiment along with qualitative and quantitative results are given in the supplementary material.

\section{Ablation Study}
\begin{figure}
  \centering
  \includegraphics[width=0.8\linewidth]{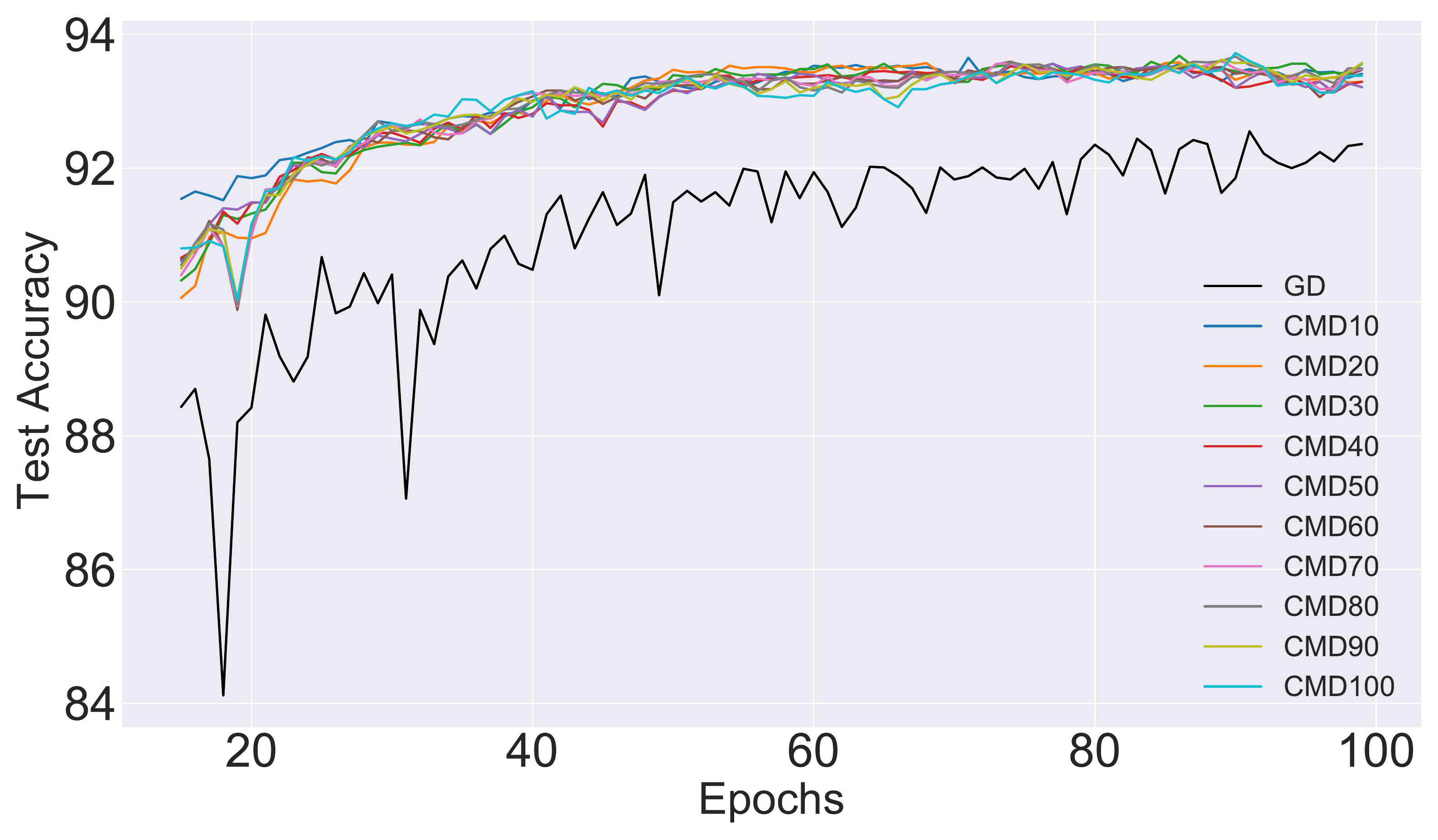}
  \caption{Robustness to number of modes. Test Accuracy for 10 CMD runs (modeling the same GD process), using different number of modes: 10-100. CIFAR10 classification, Resnet18. We can observe the model is not very sensitive to the number of modes.}
  \label{cifar_resnet18_test_acc_diff_M}
\end{figure}

\begin{figure}
  \centering
  \includegraphics[width=0.8\linewidth]{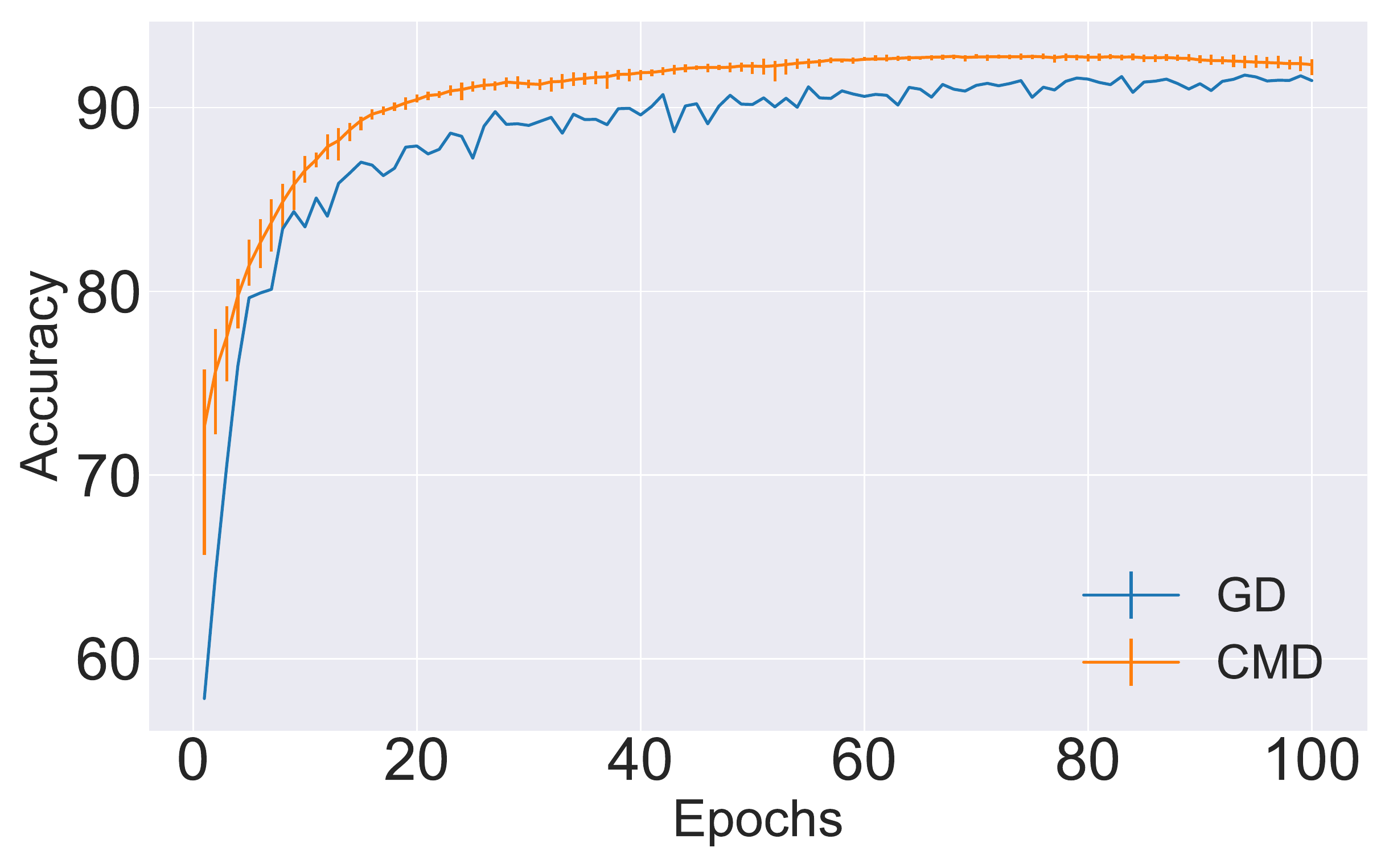}
  \caption{10 different CMD runs, K=1000 and M=10. Mean and range of Test Accuracy. Vertical line at each epoch indicates range of minimal and maximal accuracy for 10 different CMD modelings and the maximal accuracy of the original GD training. CIFAR10 Classification, Resnet18.}
  \label{10_cmd_runs_fixed_k_and_M}
\end{figure}

\begin{figure}
  \centering
  \includegraphics[width=0.8\linewidth]{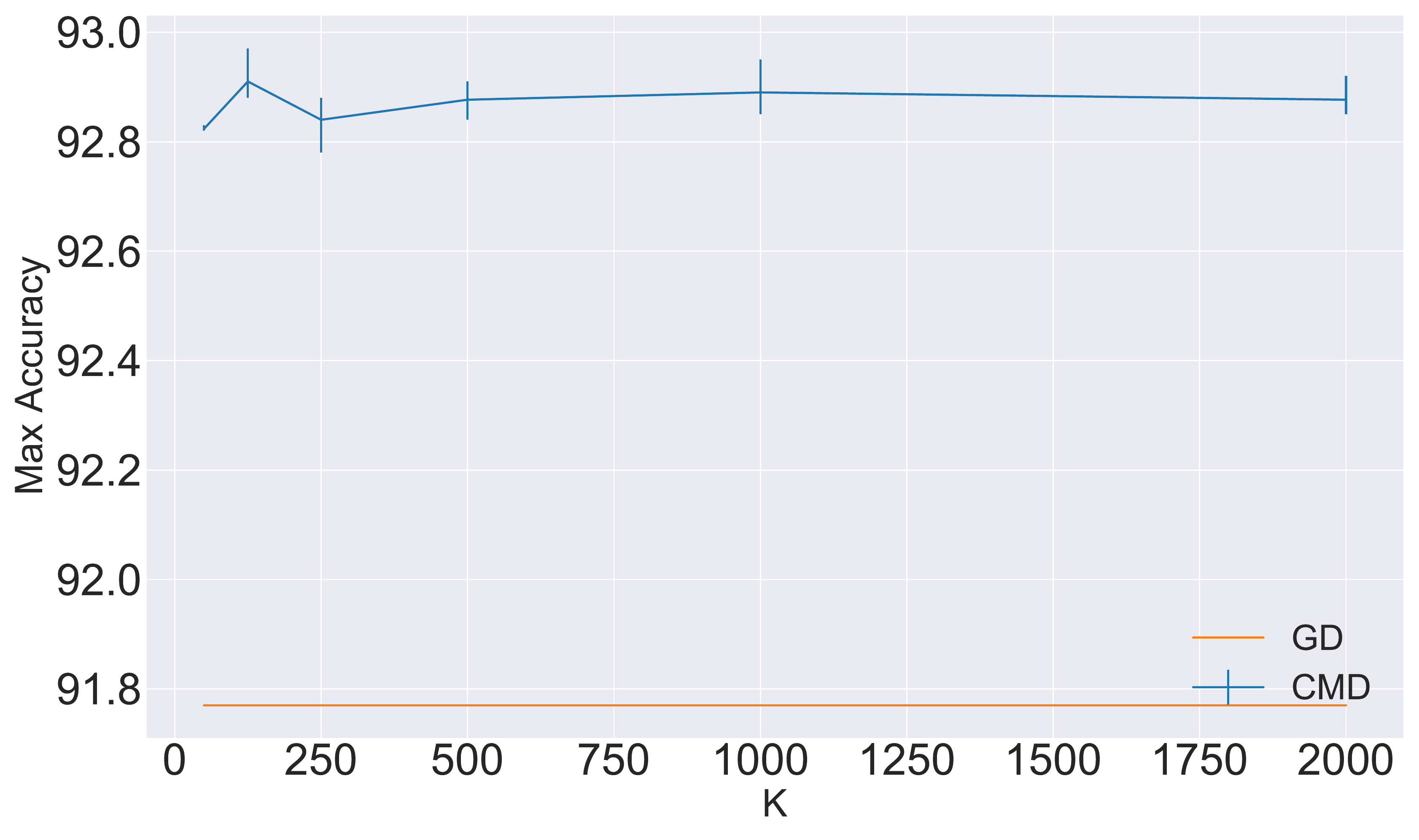}
  \caption{Mean and range of maximal Test Accuracy. Vertical line at each epoch indicates range of maximal accuracy for 5 different CMD modeling for each K value. CIFAR10 Classification, Resnet18.}
  \label{5_cmd_runs_different_k_fixed_M}
\end{figure}

\begin{figure}
  \centering
  \includegraphics[width=0.8\linewidth]{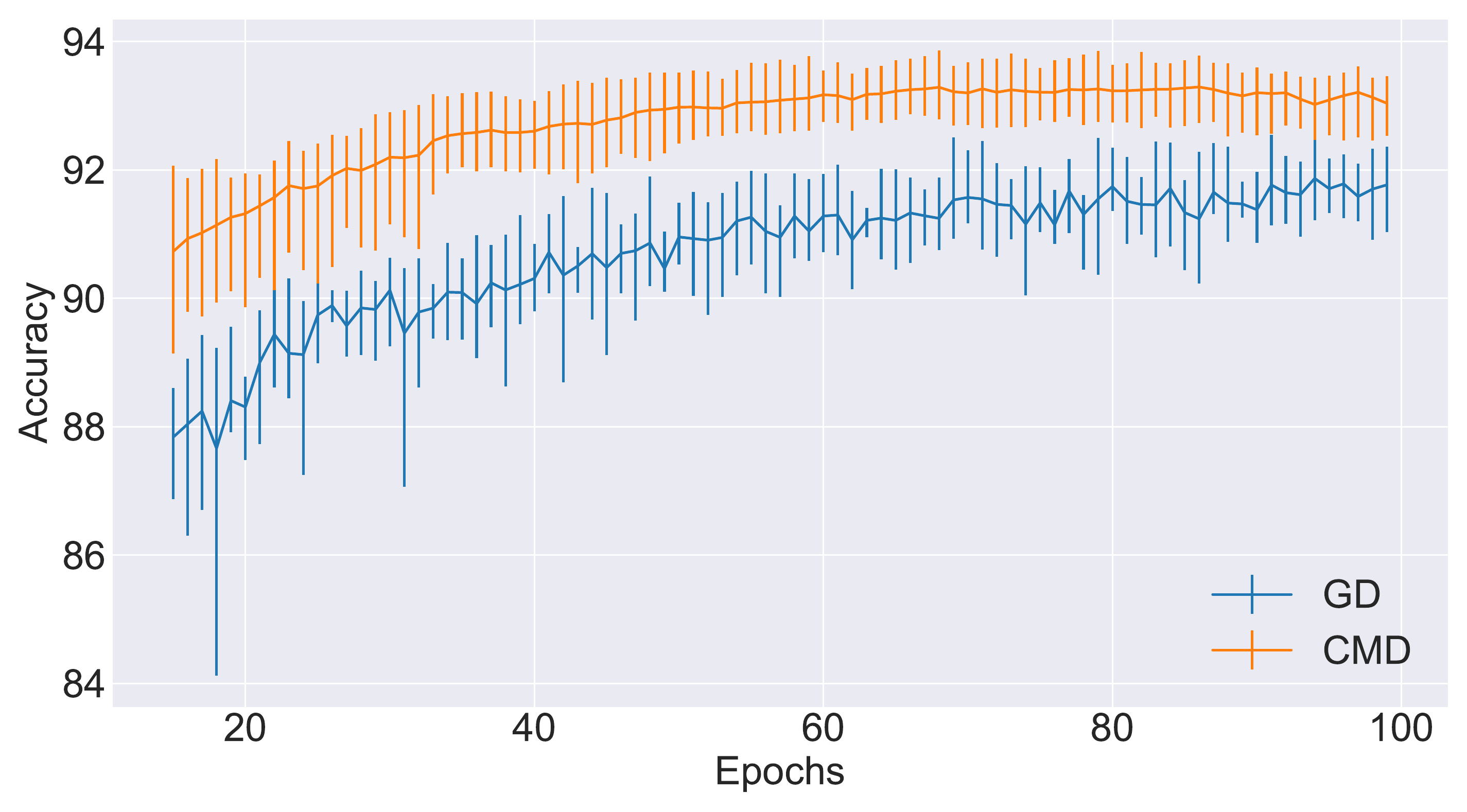}
  \caption{Mean and range of Test Accuracy. Vertical line at each epoch indicates range of minimal and maximal accuracy for 10 different GD training and their corresponding CMD modeling. CIFAR10 Classification, Resnet18.}
  \label{cifar10_resnet18_10runs}
\end{figure}

Our method introduces additional hyper-parameters such as the number of modes ($M$) and the number of sampled weights to compute the correlation matrix ($K$). We performed several robustness experiments, validating our proposed method and examining the sensitivity of the CMD algorithm to these factors.
\begin{enumerate}
    \item \textbf{Robustness to number of modes.} Fig. \ref{cifar_resnet18_test_acc_diff_M} presents the test accuracy results of 10 different CMD models, with different number of modes for the same gradient descent (GD) training. We observe only minor changes.
    \item \textbf{Robustness to different random subsets.} Here We examine empirically our assumption that in the K sampled weights we obtain all the essential modes of the network.
    In Fig. \ref{10_cmd_runs_fixed_k_and_M} we show the results of 10 CMD experiments executed on the same GD training (Resnet18, CIFAR10), each time a different random subset of the weights is sampled.
    We fixed the number of modes to $M=10$ and the subset size to $K=1000$.
    The mean value of the test-accuracy at each epoch is shown, as well as the maximal and minimal values (over 10 trials, vertical bars). 
    As can be expected, there are some changes in the CMD results. However, it is highly robust and outperforms the original GD training.
    In addition, we conducted several experiments, examining different values of $K$, the values were $K = 50, 125, 250, 500, 1000, 2000$. The number of modes was fixed ($M=10$).
    For any fixed $K$, 5 experiments were carried out, see
    Fig. \ref{5_cmd_runs_different_k_fixed_M} . We observe that here as well the algorithm is robust to this meta-parameter. The results tend to be with somewhat less variance for larger $K$, but generally we see the sampling size for the initial clustering phase can be very small, compared to the size of the net $N$.

    \item \textbf{Robustness to random initialization} - In Fig. \ref{cifar10_resnet18_10runs} we show that the effects of random initialization of the weights is marginal. The results of 10 training experiments are shown (Resnet18, CIFAR10), each time a different random initialization of the weights is used. 
\end{enumerate}

\section{Limitations}
As of now the proposed method can be applied post-training, once the weights learning trajectory is available. In principle, this requires checkpointing of the model at each epoch, which is memory consuming for large models. To deal with this drawback, we show in the supplementary material two ways of using far less memory. 
In addition, we focus in this paper on establishing a model for neural training and invest our efforts in validating the model robustness on several architectures. However we do not suggest applications yet. We assess this model can be highly instrumental in network analysis and in the design of acceleration techniques.

\section{Discussion and Conclusion}
In this paper we proposed a new approach to model the evolution of the neural network parameters through the epochs of training. We have shown that there are essentially only a few typical evolution profiles, which can model well the network dynamics. Full fledged networks, containing millions of parameters, can be well represented by ten or less modes. We also observed that although one may expect that the CMD approximation yields reduced performance, our model induces regularization of the training, allowing better generalization capacity and improved accuracy results on the test set.
This can be related to earlier studies which show that weighted averaging yields better minima, improving generalization capacity \cite{izmailov2018averaging,tarvainen2017mean}. Those methods, however, do not yield a low dimensional model of the network. Our method is fully data-driven and has no implicit assumptions on the learning rate or the specific underlying optimizer (for instance, it works well with ADAM \cite{kingma2014adam}). 

There are many open questions on how to analytically justify the model and its dependence on the architecture, methods of training and type of training set. We believe our findings can pave ways to improve the understanding of the training process of neural networks and to allow the design of better and faster optimization schemes.

{\small
\bibliographystyle{unsrt}
\bibliography{smartPeople.bib}
}
\newpage

\section{Supplementary Material}
This Appendix provides supplementary details and results.

\subsection{Memory saving}
\begin{figure}[H]
     \centering
     \includegraphics[width=\linewidth]{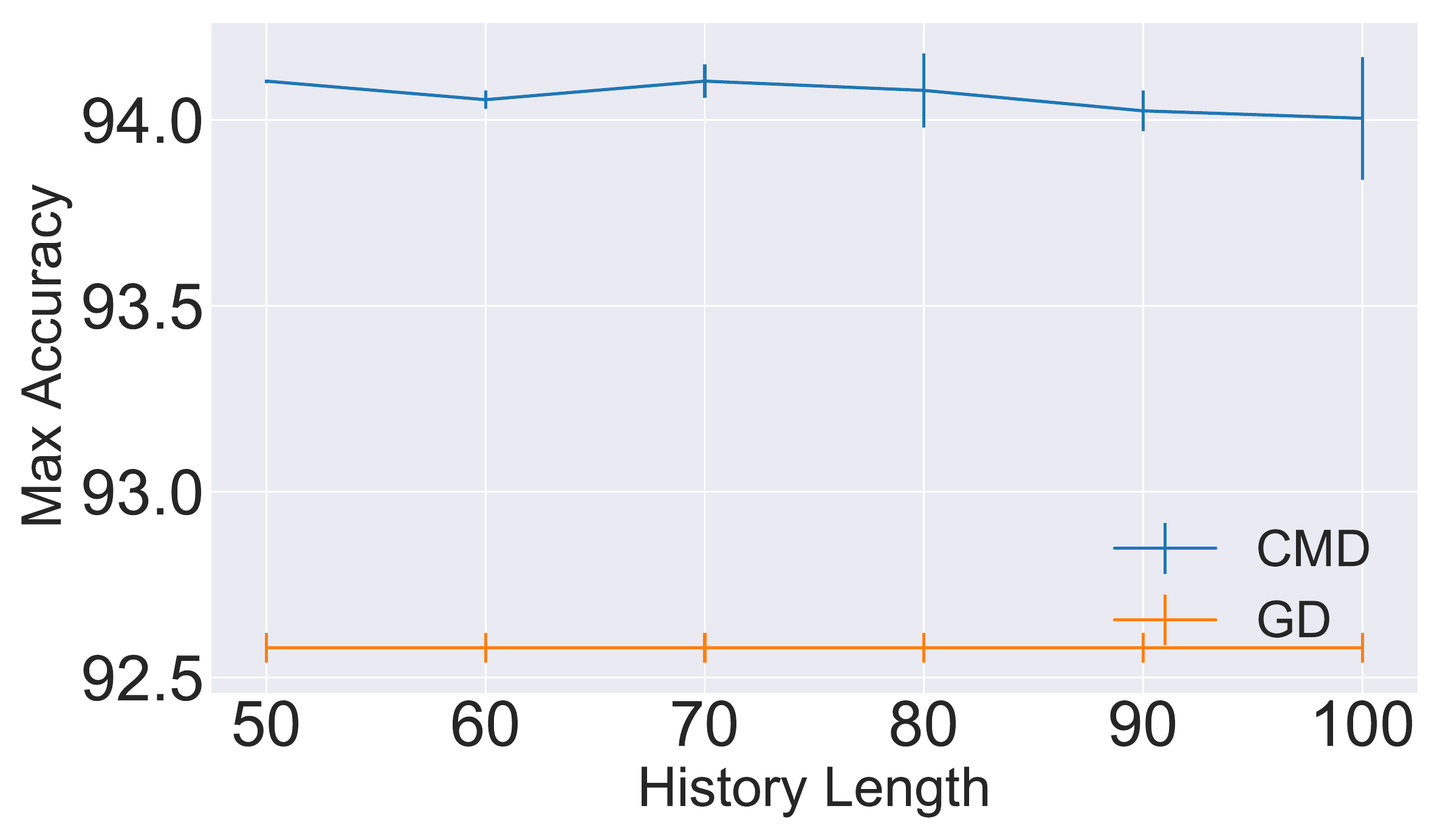}
     \caption{CMD modeling Vs. history length CIFAR10 Classification, Resnet18.}
     \label{different_history_length}
    \end{figure}

\begin{figure}[H]
      \centering
      \includegraphics[width=\linewidth]{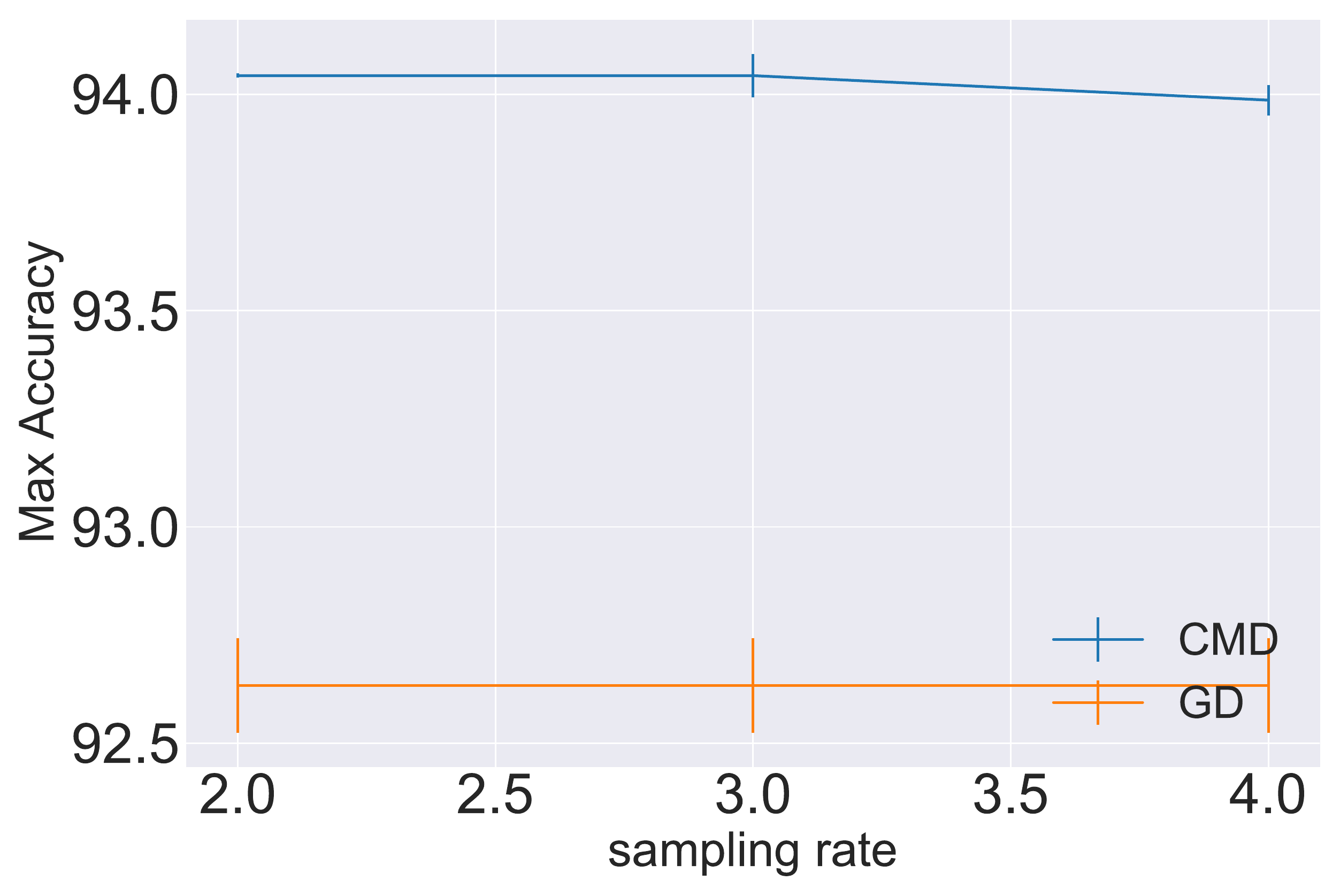}
      \caption{CMD modeling Vs. sampling rates of checkpoints. CIFAR10 Classification, Resnet18.}
      \label{different_sampling_rates}
    \end{figure}

The method we suggest in this paper is applied post-training, and requires checkpointing the entire model at each epoch. This, naturally, is memory intensive and may be challenging for very large models. In this paper we focused on the general model and novel observations and not on memory consumption. However we would like to suggest two preliminary options to save memory. We show that the model parameters can be estimated well using fewer checkpoints than all epochs. As described in the main paper, there are two main stages of the CMD algorithm - first we calculate correlations and perform clustering based on a very limited number of parameters (in all our experiments $K=1000$). For this stage memory consumption is negligible. The second part, after basic clustering and reference weights $w_r$ are found, is clustering the rest of the network parameters and computing the affine coefficients $a,b$. This may be done more sparingly, without storing the entire training history. Two ways are suggested:
\begin{enumerate}
    \item \textbf{Storing shorter history} - here we suggest to start storing the training checkpoints only from later epochs. In order to test this option, we executed different experiments, where we examine the affect of using different length of training history on the performance of our method. The training history lengths that were tested are: $100 (full), 90, 80, 70, 60, 50$. The results presented in Fig. \ref{different_history_length} show that as less history (from the last epoch) is taken into account the model actually behaves better, with less variance. This may be due to the fact that this phase of the training is more stable. 

    \item \textbf{Sub-sampling of the epochs} - here we suggest to sub-sample the epoch space by an integer factor. For simplicity, we check na\"ive sub-sampling, where no low-pass filtering is performed beforehand (hence aliasing may occur) and check sub-sampling by factors of 2, 3 and 4 (store only every 2nd epoch, 3rd epoch or 4th epoch). This reduces the size of the checkpoints by the sub-sampling factor. The results of these experiments are shown in Fig. \ref{different_sampling_rates}. We see that the model behaves well, with only a very minor decrease in performance as the ratio increases.
    
\end{enumerate}

\subsection{Sampling Details}
We would like to make sure that every layer is represented in the subset that is randomly sampled at the first step of our method (correlation and clustering). The choice of reference weights depends on this set. We therefore allocate a budget of $\frac{K}{2\cdot|layers|}$ for each layer and randomly sample it uniformly (using a total budget of $\frac{K}{2}$ parameters). The other $\frac{K}{2}$ parameters are sampled uniformly throughout the network parmeter space. In case where $\frac{K}{2}<=|layers|$ we simply sample all of the $K$ parameters uniformly (as done in section 1.3, where we examined smaller values of $K$).

\subsection{Clustering Method}
To cluster the correlation map into modes we employed a full linkage algorithm - the Farthest Point hierarchical clustering method. After creating a linkage graph we broke it down into groups either by thresholding the distance between vertices (parameter $t$), or by specifying a number for modes ($M$), as depicted in section 3.2 in the main paper (\emph{CMD Algorithm}). These two steps, however, could be substituted by a simpler clustering algorithm such as k-means and others. 

\subsection{Experimenting on a ViT-b-16 transformer}
Here we show that generally similar properties hold for an architecture which is significantly different than Resnet18.

\begin{figure}[H]
  \centering
  \includegraphics[width=\linewidth]{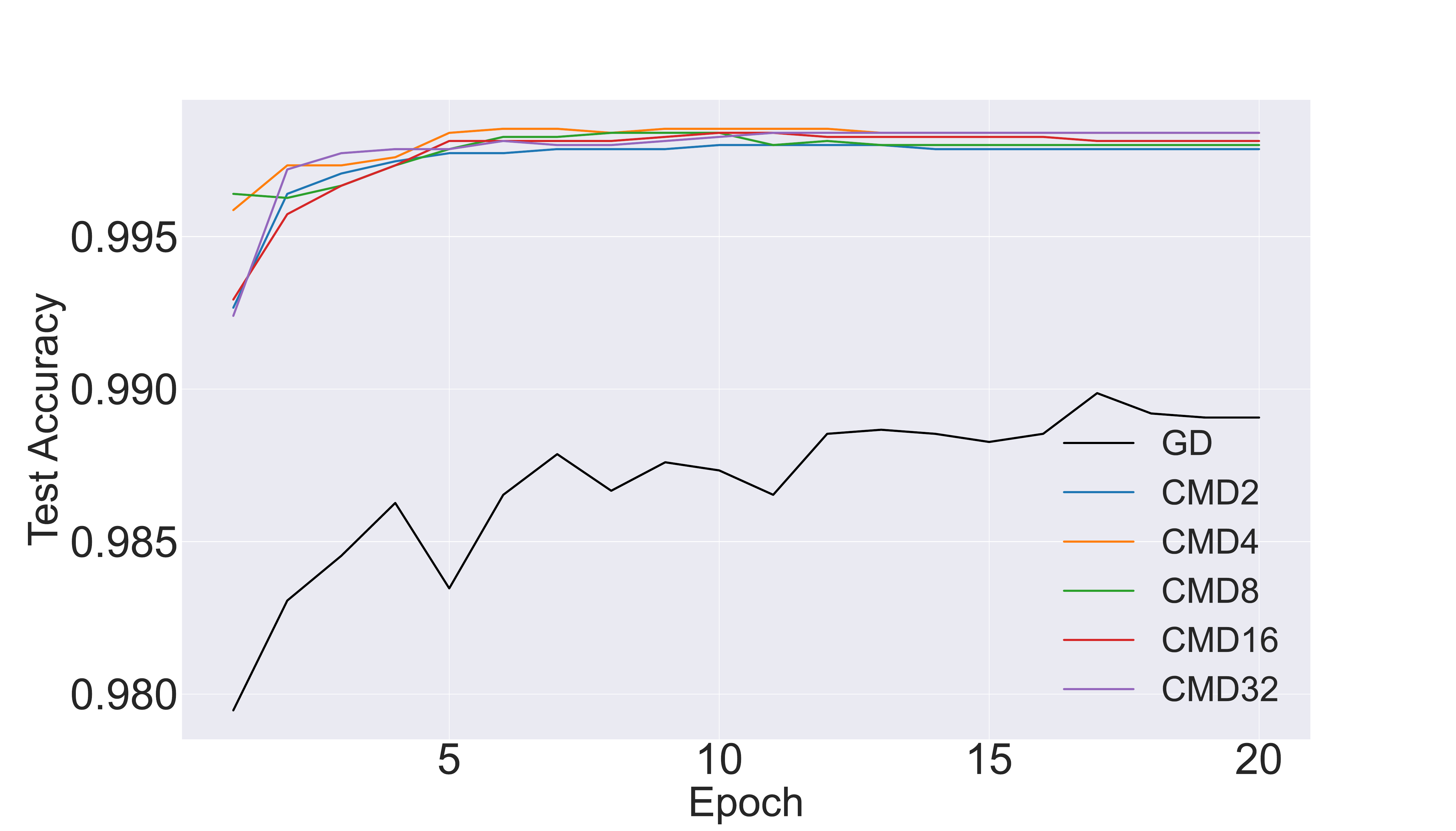}
  \caption{Test Accuracy for 5 CMD models, with different number of modes: $2^1 \dots 2^5$ vs. GD. CIFAR10 classification, pre-trained ViT-b-16 transformer. We observe the model is almost insensitive to the number of modes used.
  }
  \label{vit_pretrained_test_acc_diff_M}
\end{figure}

\begin{figure}[H]
  \centering
  \includegraphics[width=\linewidth]{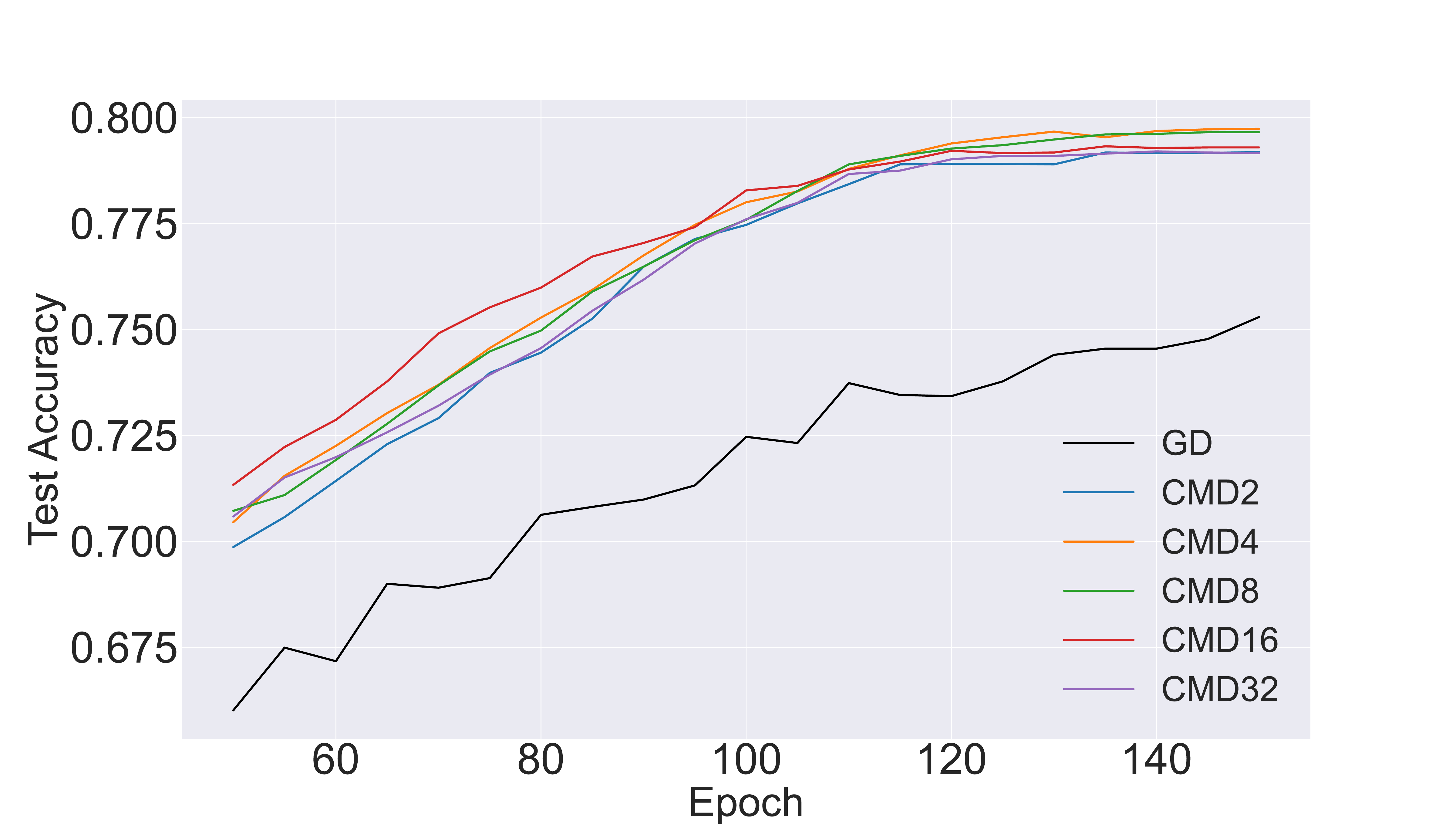}
  \caption{Test Accuracy for 5 CMD models, with different number of modes: $2^1 \dots 2^5$ vs. GD. CIFAR10 classification, ViT-b-16 randomly initialized transformer (training from scratch). The model shows robustness to number of modes by performing similarly well in all configuarations.
  }
  \label{vit_xavier_test_acc_diff_M}
\end{figure}

We ran a new experiment fine-tuning a pre-trained ViT-b-16 transformer on CIFAR10, this time with a larger batch size of 32 (vs. 8 in the experiment depicted in section 4 in the paper). We apply CMD with a varying number of modes $2^1 \dots 2^5$ on the fine-tune training process. The results are shown in figure \ref{vit_pretrained_test_acc_diff_M}. we still achieve good performance in every configuration despite the fact that the GD test accuracy only reached 98.7\% this time. 
In addition, we repeat the above experiment for a randomly initialized ViT-b-16 transformer, thus training it anew on CIFAR10, with batch size of 75. We plot the results of the original GD training test accuracy vs. CMD modelling in figure \ref{vit_xavier_test_acc_diff_M}. The modelling was done from epoch 50 to epoch 150 and the results are sampled in gaps of five epochs. 

\subsection{Image synthesis for multiple domains using StarGan-V2 on CelebHQ}
In this section we evaluate our method on image synthesis task. We used the StarGAN v2 framework developed in \cite{choi2020stargan}.

This framework consists of four modules that were jointly trained in the same way described in \cite{choi2020stargan}, using the code provided at \href{https://github.com/clovaai/stargan-v2}{\color{blue}{clovaai/stargan-v2}}. We preformed the CMD algorithm on each network independently. Each one of them consists of millions parameters that were modeled by a few dozens. In this experiment, the number of modes for each network was chosen with the "In-cluster distance threshold" criterion, where $t$ is half of the maximal distance between pairs.
The CMD algorithm was executed only on the generator, the mapping network and the style-encoder. Each one of these networks contains millions of parameters that were modeled by a few dozens: 
\begin{enumerate}
    \item Generator - contains 43.5M parameters that were modeled with 20 modes.
    \item Mapping network - contains 24M parameters that were modeled with 8 modes.
    \item Style encoder - 21M parameters that were modeled with 40 modes.
\end{enumerate}

We used this framework to synthesize images in two different ways:
\begin{itemize}
    \item Latent-guided synthesis: The style code given to the generator is produced by the mapping network, that learned to translate a random latent code to style code in the domains it was trained on. In these experiments the mapping network was trained to generate styles for two domains - male and female.
    \item Reference-guided synthesis: The style code given to the generator is produced by the style- encoder network that was trained to extract style from a given image. In these experiments we synthesized images of a given image (source) when the style was taken from another image (reference).
\end{itemize}

\begin{figure}
  \centering
  \includegraphics[width=0.9\linewidth]{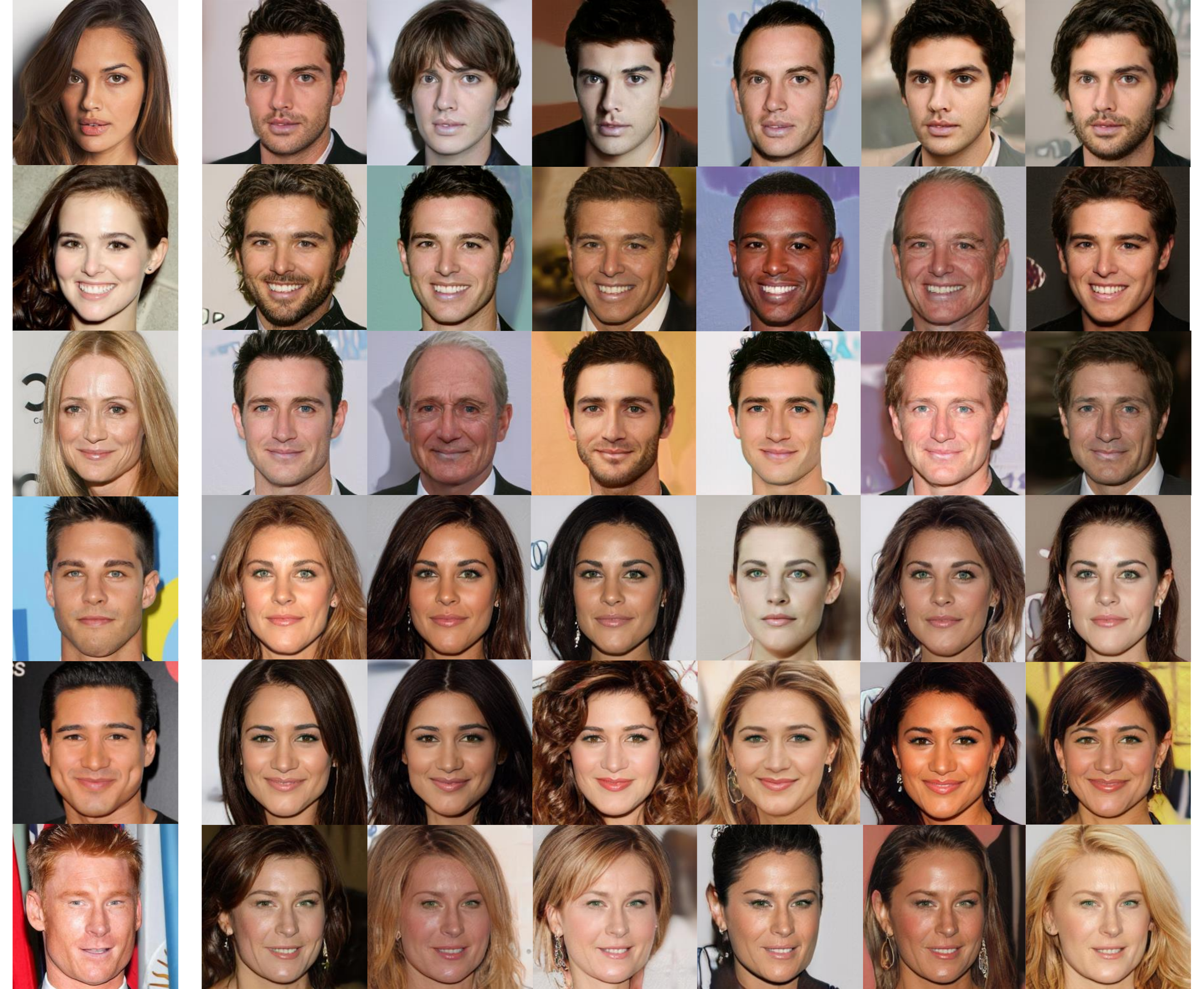}
  \caption{Latent-guided image synthesis results on CelebA-HQ. In the left-most column are the source images that were translated to target domains with our CMD models using randomly sampled latent codes.}
  \label{latent_guided_res}
\end{figure}

\begin{figure}
  \centering
  \includegraphics[width=0.9\linewidth]{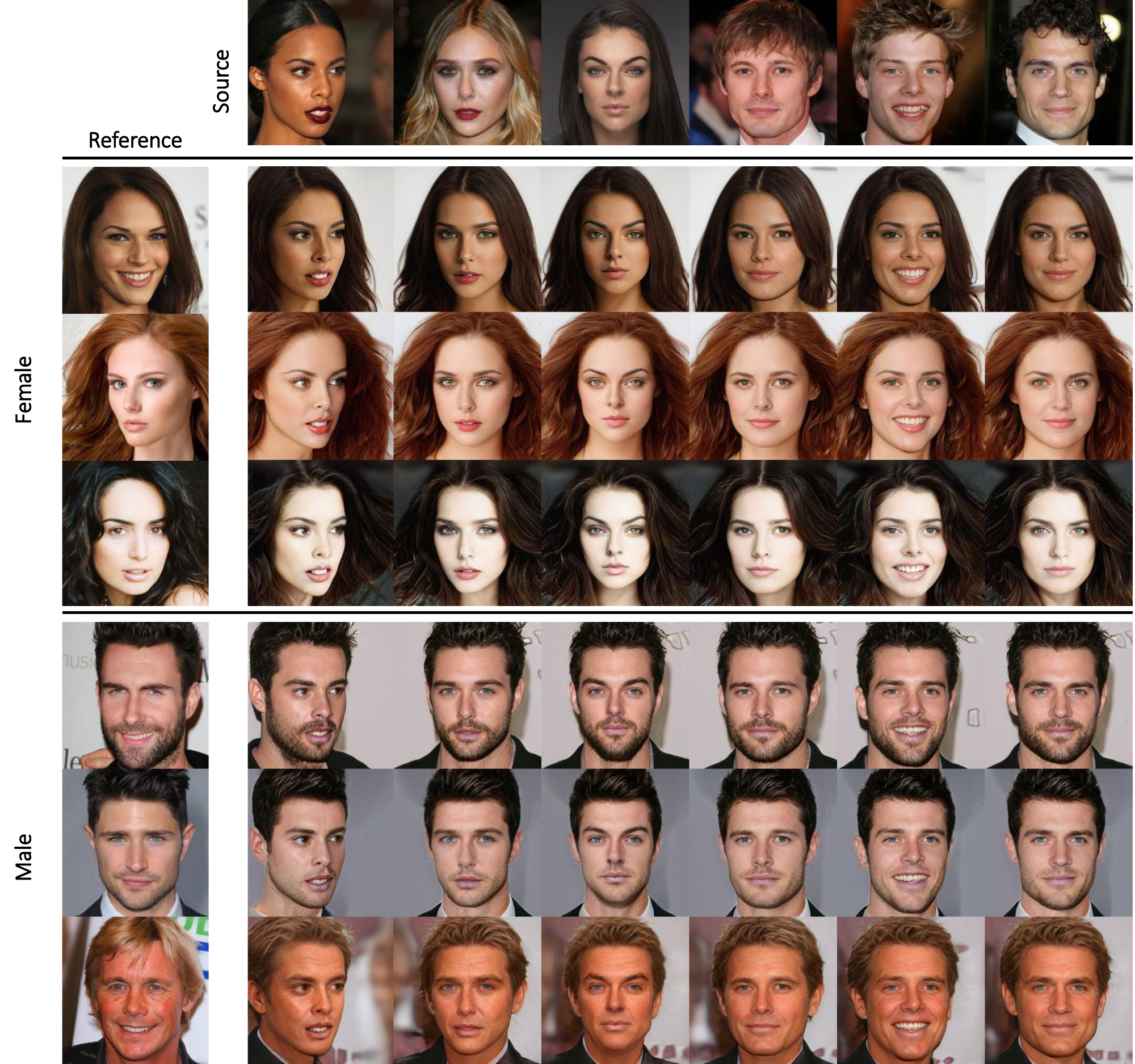}
  \caption{Reference-guided image synthesis results on CelebA-HQ. The source and reference images in the first row and the first column are real images, while the rest are images generated by the models created after applying our CMD method on the original snapshots.}
  \label{reference_guided_res}
\end{figure}

\begin{figure*}[t]
  \centering
  \includegraphics[width=\linewidth]{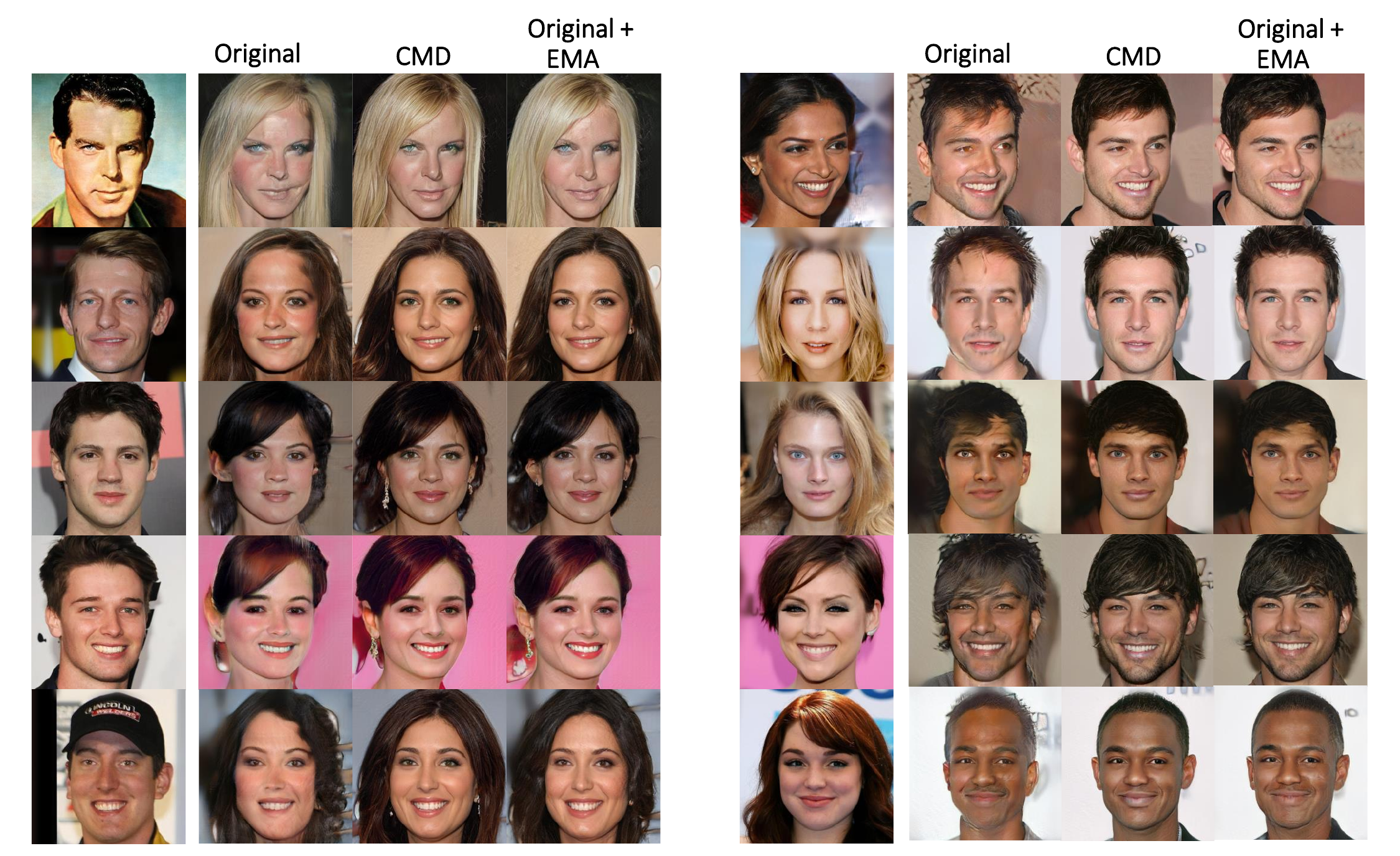}
  \caption{Qualitative comparison of latent-guided image synthesis results on CelebA-HQ. The source images in the first column are real images, while the rest are the translation to target domains generated by the original trained networks, our CMD models and the original models after applying EMA, respectively. (\textbf{Left}) Source domain: male, Target domain: female. (\textbf{Right}) Source domain: female, Target domain: male.}
  \label{latent_guided_res_and_comp}
\end{figure*}

\begin{figure*}
  \centering
  \includegraphics[height=135mm]{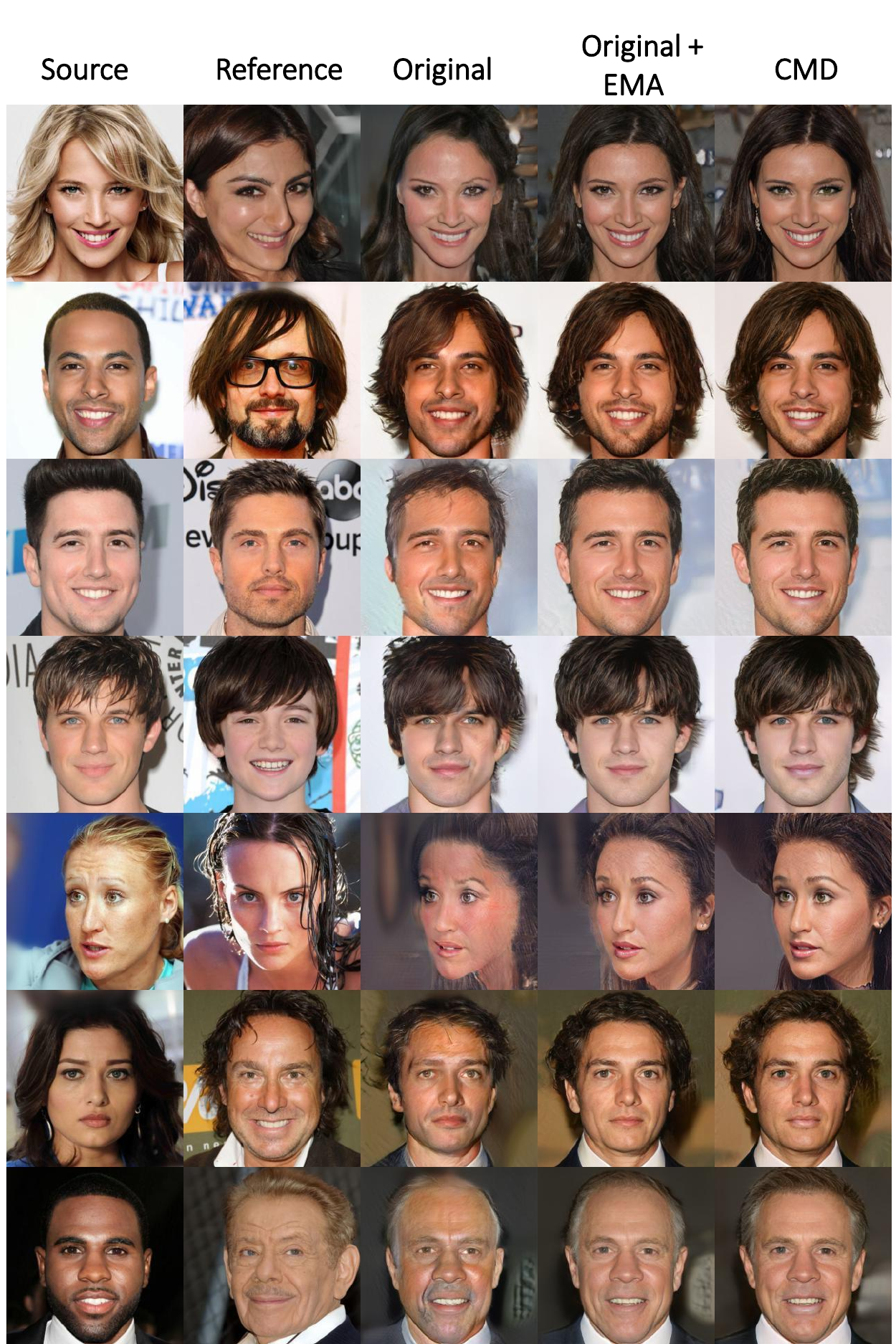}
  \caption{Qualitative comparison of reference-guided image synthesis results on CelebA-HQ. The source and reference images in the first two columns are real images, while the rest are images generated by the original trained networks, the original EMA networks and the models created by applying our CMD method on the original snapshots, respectively.}
  \label{reference_guided_comp_to_ema}
\end{figure*}

\begin{figure*}
  \centering
  \begin{subfigure}{.5\linewidth}
  \centering
  \includegraphics[width=\linewidth,height=70mm]{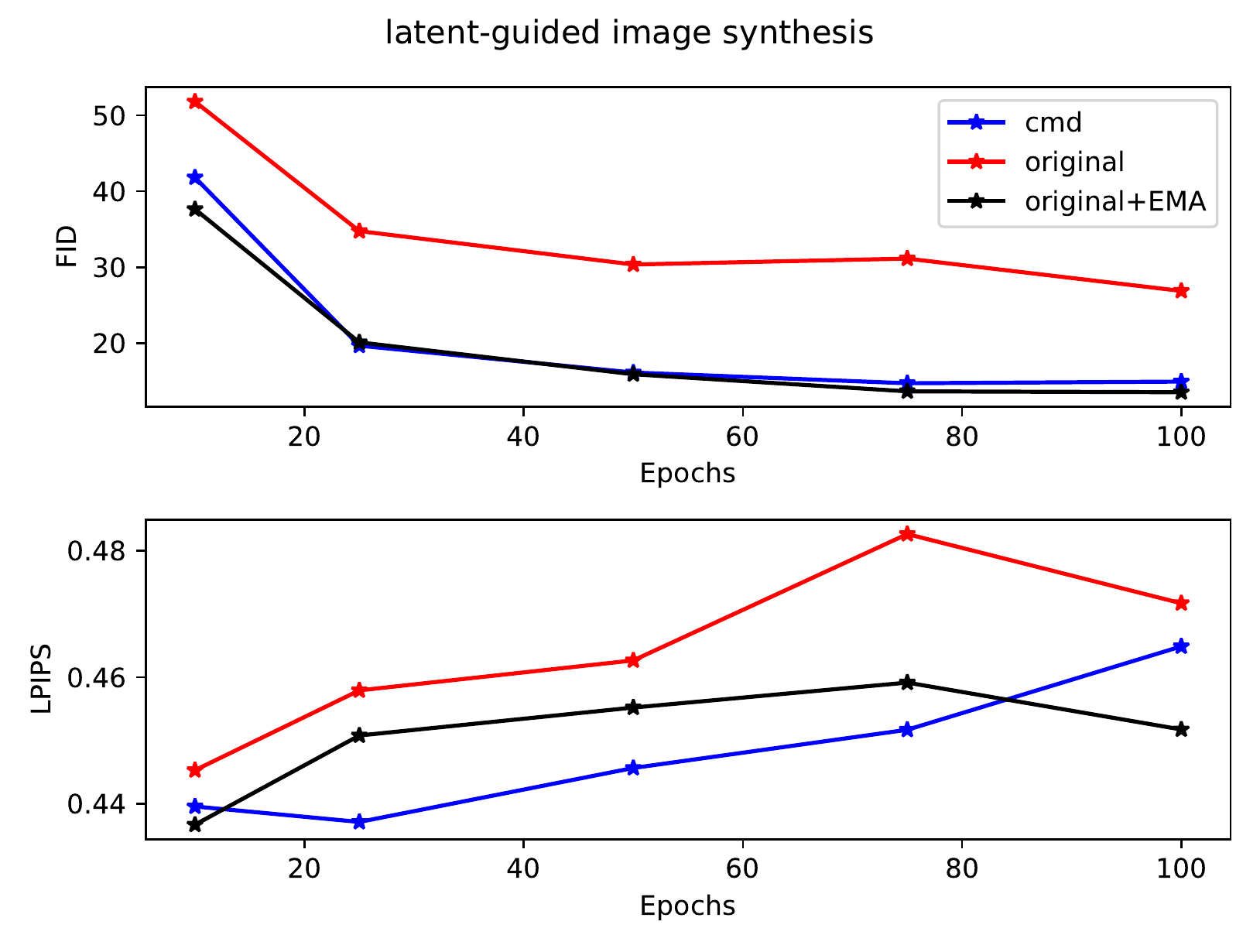}
  \caption{Latent-guided image synthesis}
  \label{stargan_latent_fid_lpips_res:sub1}
  \end{subfigure}
  \begin{subfigure}{.49\linewidth}
  \centering
  \includegraphics[width=\linewidth,height=70mm]{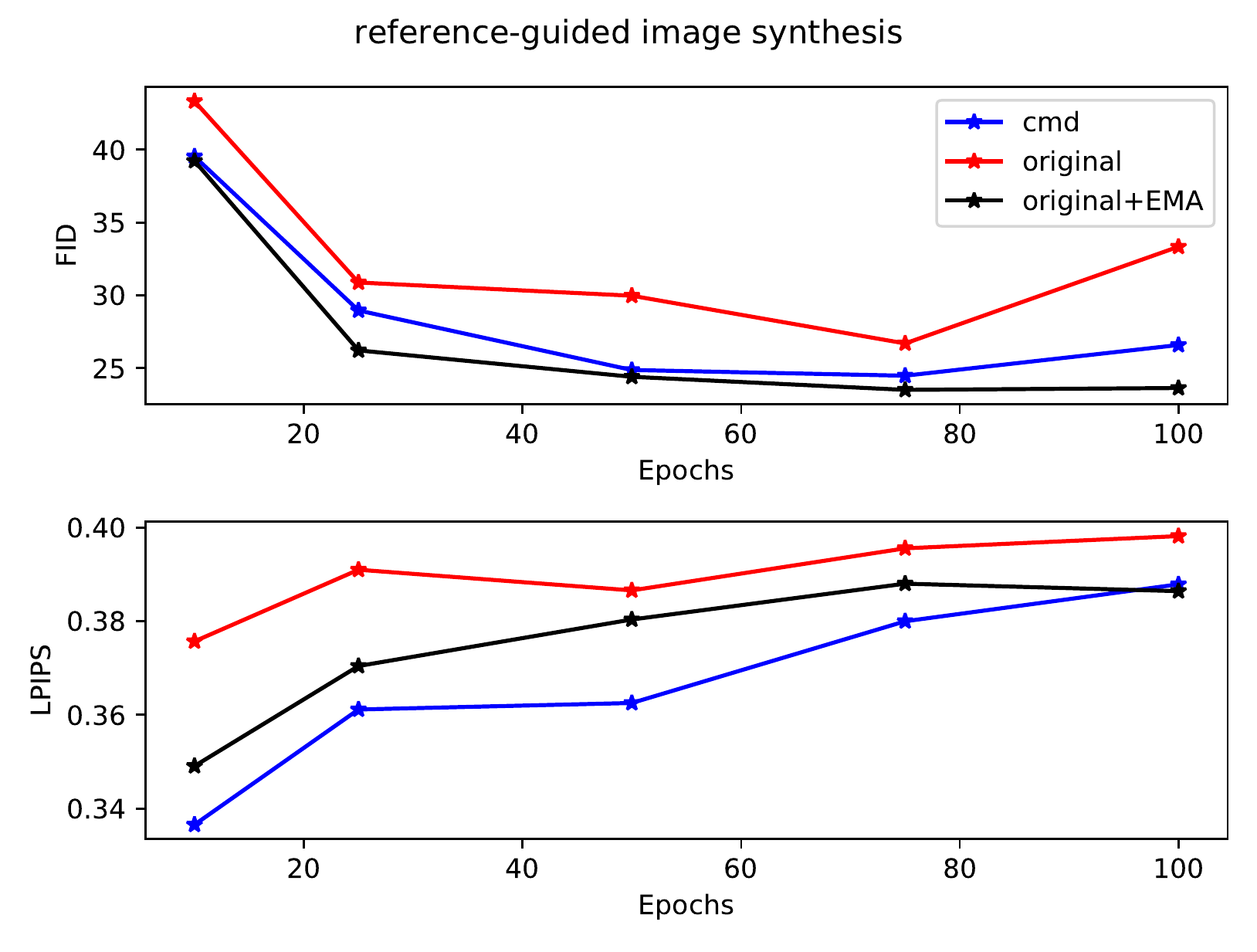}
  \caption{Reference-guided image synthesis}
  \label{stargan_latent_fid_lpips_res:sub2}
  \end{subfigure}
  \caption{StarGan-v2, CelebHQ. Quantitative comparison between original Gradient Descent (GD) training, EMA models and CMD modeling. (a) Latent-guided synthesis. (b) Reference-guided synthesis. Up: FID, Down: LPIPS.
  }
  \label{stargan_latent_fid_lpips_res}
\end{figure*}

We present in figures \ref{latent_guided_res}, \ref{reference_guided_res} the performance of the "CMD models" for latent-guided image synthesis and reference-guided image synthesis on CelebA-HQ. In the original paper Stargan-v2 was evaluated after employing exponential moving averages over the parameters of all modules except the discriminator. We executed the CMD over the original nets (without EMA), and compared our results to the original GD training and to the original networks after EMA is applied.
A qualitative comparison is presented in figures \ref{latent_guided_res_and_comp} \ref{reference_guided_comp_to_ema}.
In addition, we evaluated the results using the FID and LPIPS metrics. A quantitative comparison is presented in Fig. \ref{stargan_latent_fid_lpips_res}. One can see that our method outperforms the original GD training, and is comparable to the EMA models - a reasonable result, since we have noticed that our method has some smoothing effect on the weights.
Thus, on such a set of complex networks we are able to recover both GD and the EMA effect through our model.

\subsection{DMD failure in modeling complex architectures}
\begin{figure*}
  \centering
  \includegraphics[height=70mm,width=0.6\linewidth]{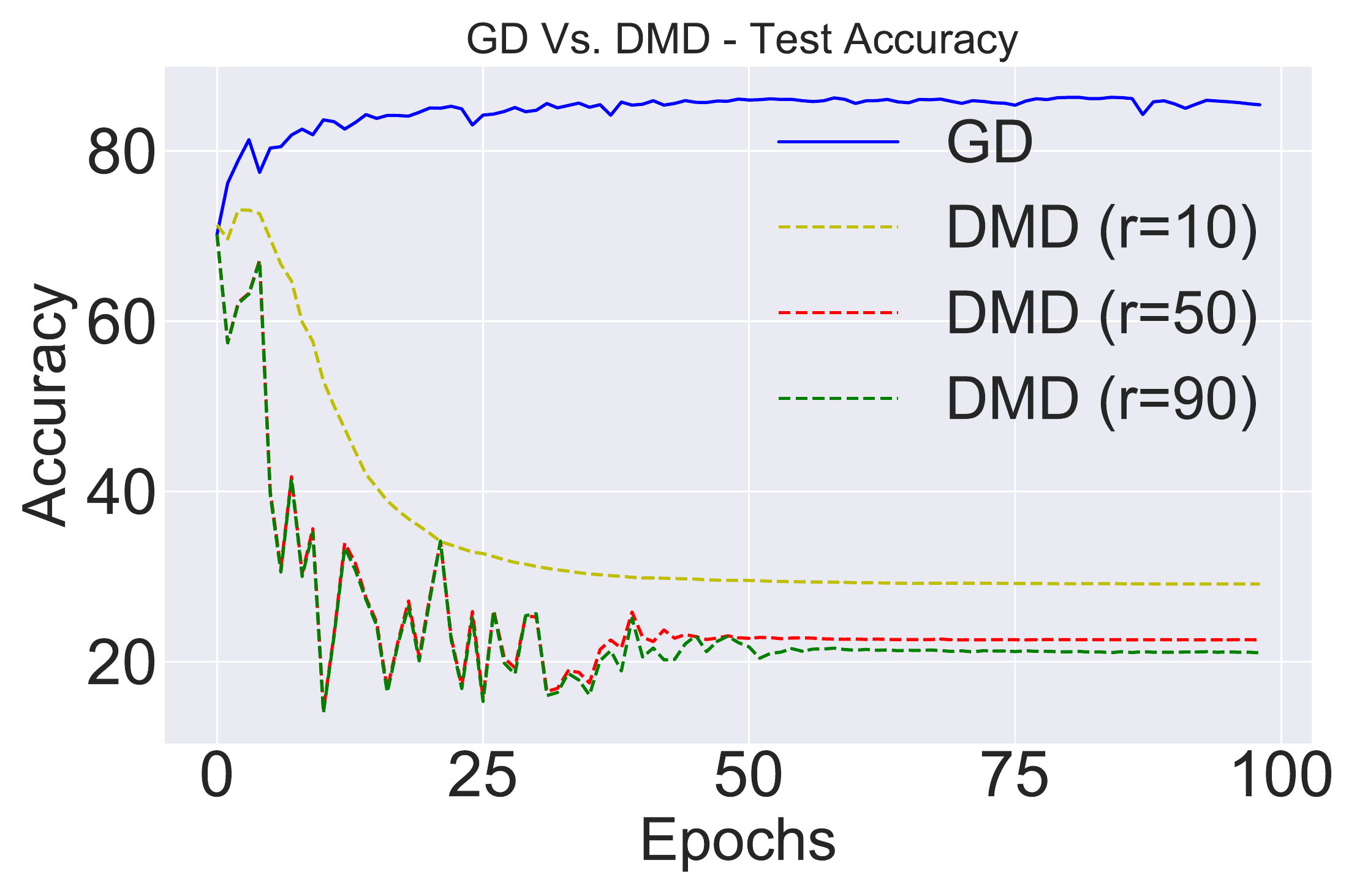}
  \caption[DMD failure with complex architectures.]{DMD failure on CIFAR10 classification using Resnet18. Test accuracy of DMD modeling for 3 values of dimensionality reduction parameter $r$.}
  \label{DMD_failure_resnet18}
\end{figure*}

In many cases where the number of the network's parameters increases, the full network DMD operator is impossible to extract for computational reasons, and the sub-system approximation as suggested in \cite{dogra2020optimizing} is used (Node koopman operator, layer koopman operator, etc.).
The problem is that this approximation does not always hold, and in cases where the architecture is more complex, the DMD fails to capture the real dynamics of the NN weights. 
We demonstrate this flaw by reconstructing Resnet18 weights with node koopman operator.
This network was trained without augmentations, in order to decrease  non-linear effects on the process. Then, we grouped weights and biases by the nodes they are connected to, and reconstructed the DMD operator for each, independently.
We ran this experiment for 3 different dimensionality reduction rates, $r=10,50,90$.
Figure \ref{DMD_failure_resnet18} presents the severe breakdown of the DMD reconstruction in this case. we believe that due to the complex structure (skip connections) of the Resnet model, the node koopman operator approximation is not suitable.

\subsection{Visualization of modes and reference weights}
\begin{figure}
  \centering
  \begin{subfigure}{\textwidth}
  \centering
   \includegraphics[width=\linewidth,height=30mm]{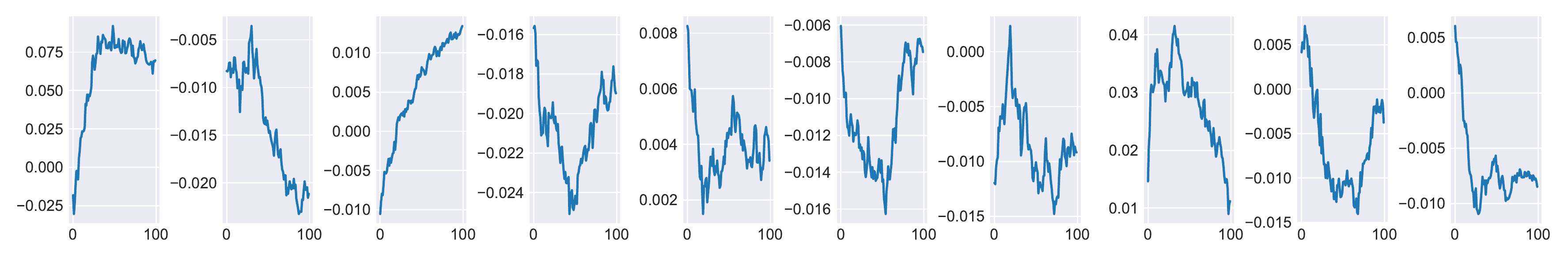}
  \caption{$w_r$ of CMD10}
  \label{cifar10_resnet18_modes:sub1}
  \end{subfigure}\\
  \begin{subfigure}{.5\textwidth}
  \centering
  \includegraphics[width=\linewidth,,height=30mm]{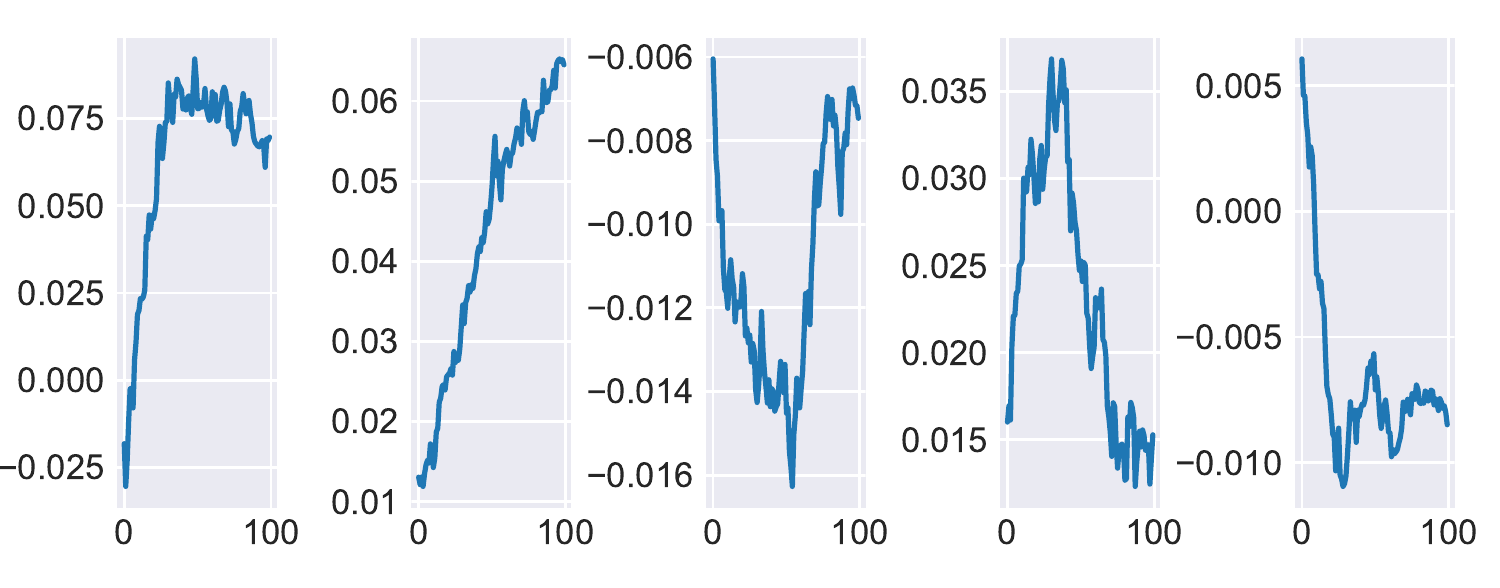}
  \caption{$w_r$ of CMD5}
  \label{cifar10_resnet18_modes:sub2}
  \end{subfigure}\hspace{10mm}
  \begin{subfigure}{.22\textwidth}
  \centering
  \includegraphics[width=\linewidth,,height=30mm]{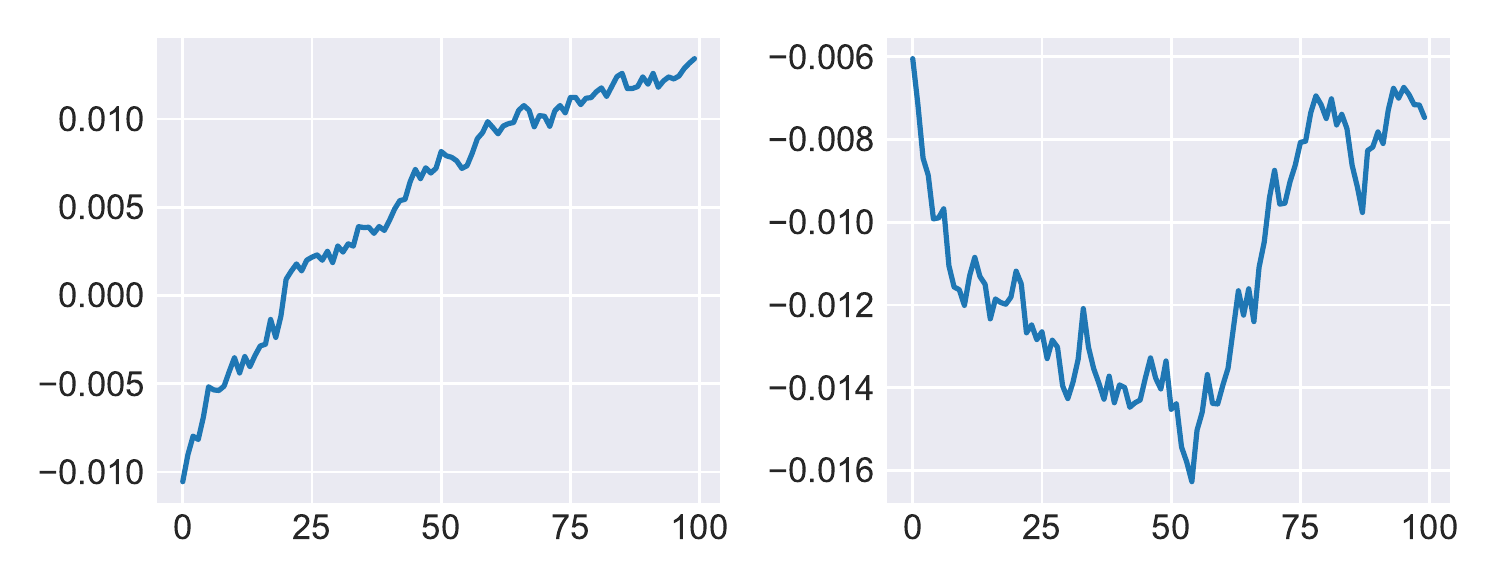}
  \caption{$w_r$ of CMD2}
  \label{cifar10_resnet18_modes:sub3}
  \end{subfigure}
  \caption{The reference weights of the CMD extracted for Resnet18, CIFAR10 classification.\\ (a) 10 modes CMD. (b) 5 modes CMD. (c) 2 modes CMD (correlation of 0.22 between the modes).}
  \label{cifar10_resnet18_modes}
\end{figure}

The dynamics of the reference weights for two, five and ten modes in the CIFAR10 classification with Resnet18 is presented in Fig. \ref{cifar10_resnet18_modes}.

\begin{figure}
  \centering
  \begin{subfigure}{.5\textwidth}
  \centering
  \includegraphics[width=\linewidth, height=37mm]{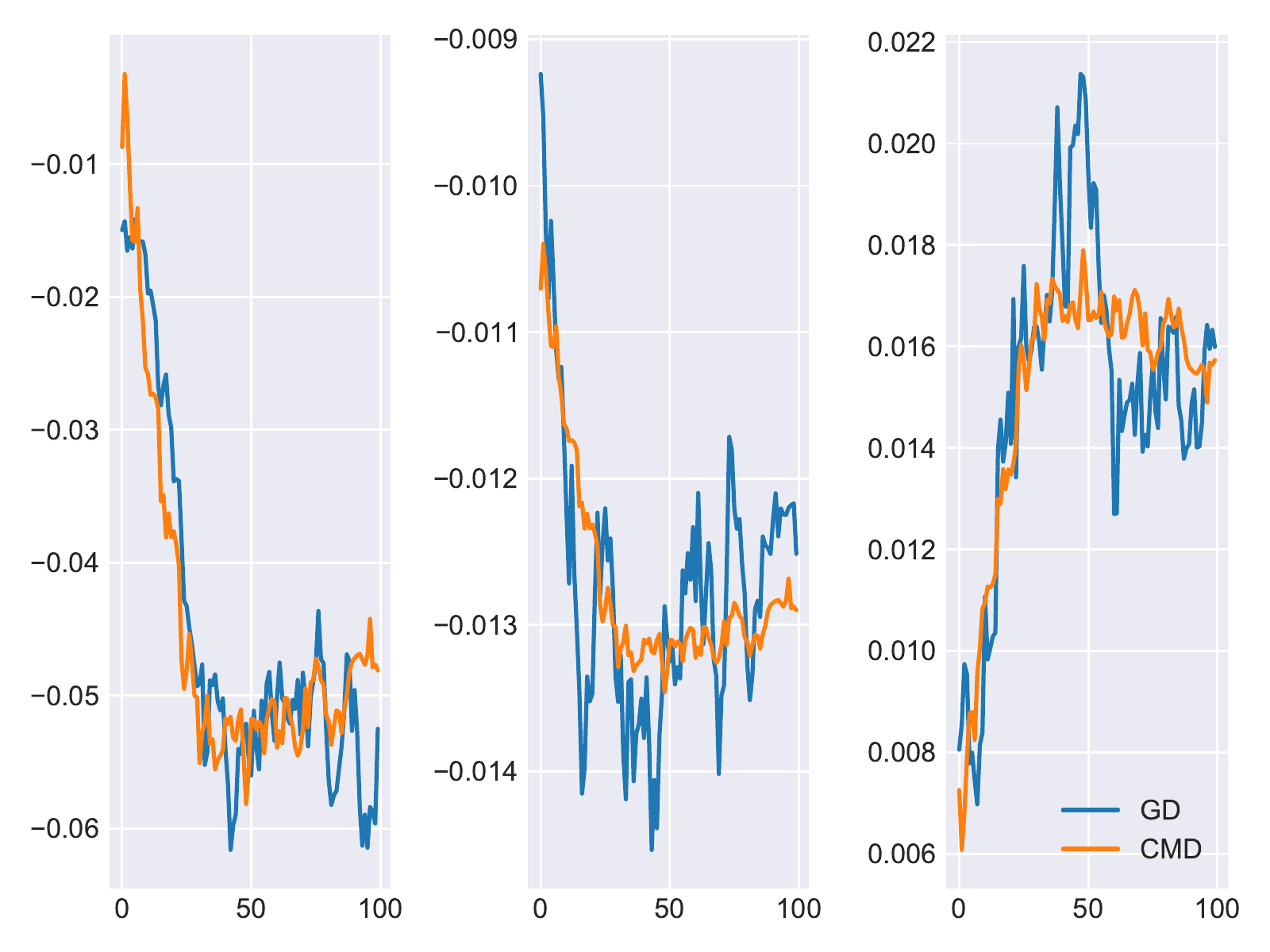}
  \caption{mode 0}
  \label{w_rec:sub1}
  \end{subfigure}
  \begin{subfigure}{.49\textwidth}
  \centering
  \includegraphics[width=\linewidth, height=37mm]{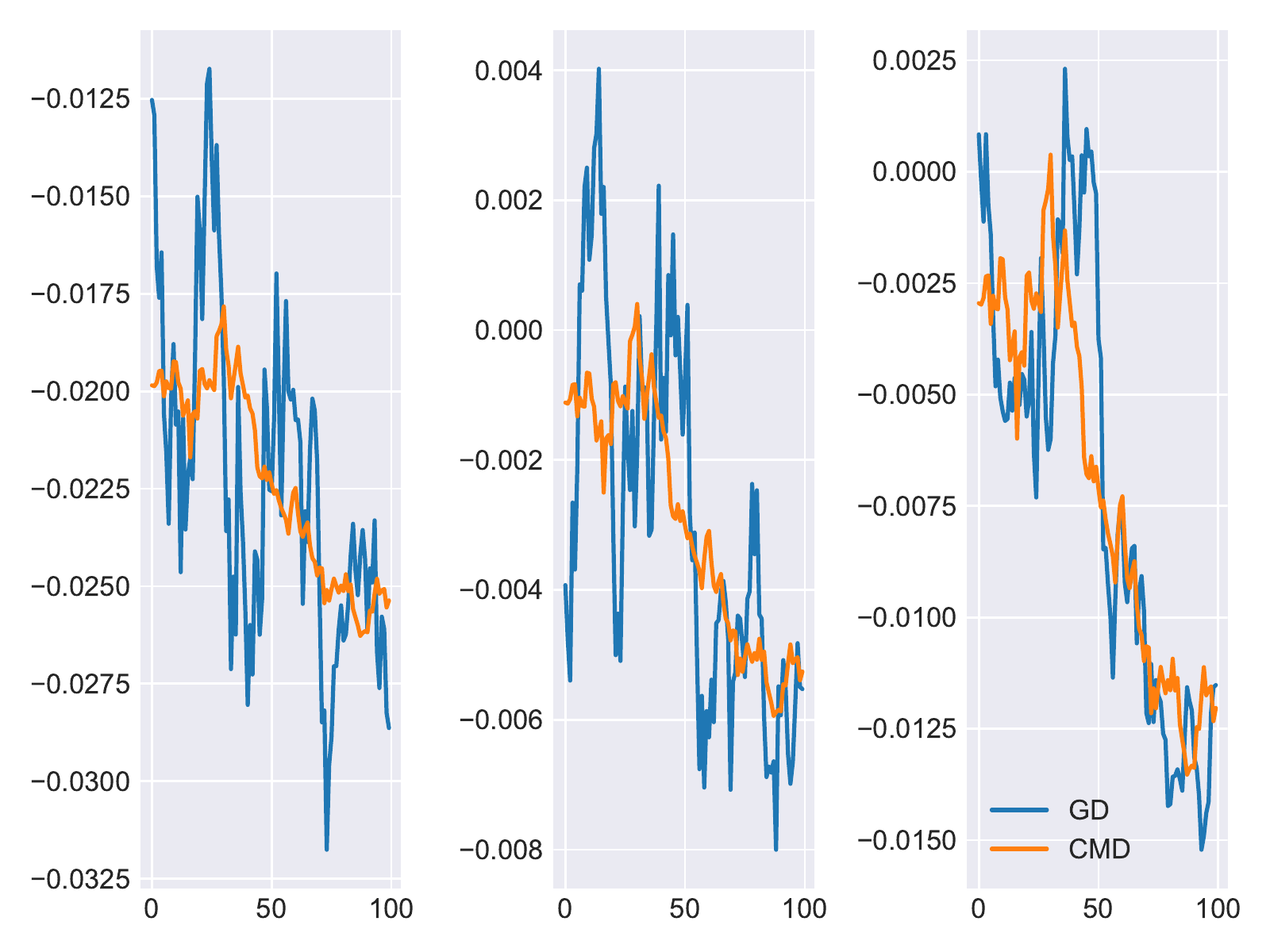}
  \caption{mode 1}
  \label{w_rec:sub2}
  \end{subfigure}
  \begin{subfigure}{.5\textwidth}
  \centering
  \includegraphics[width=\linewidth, height=37mm]{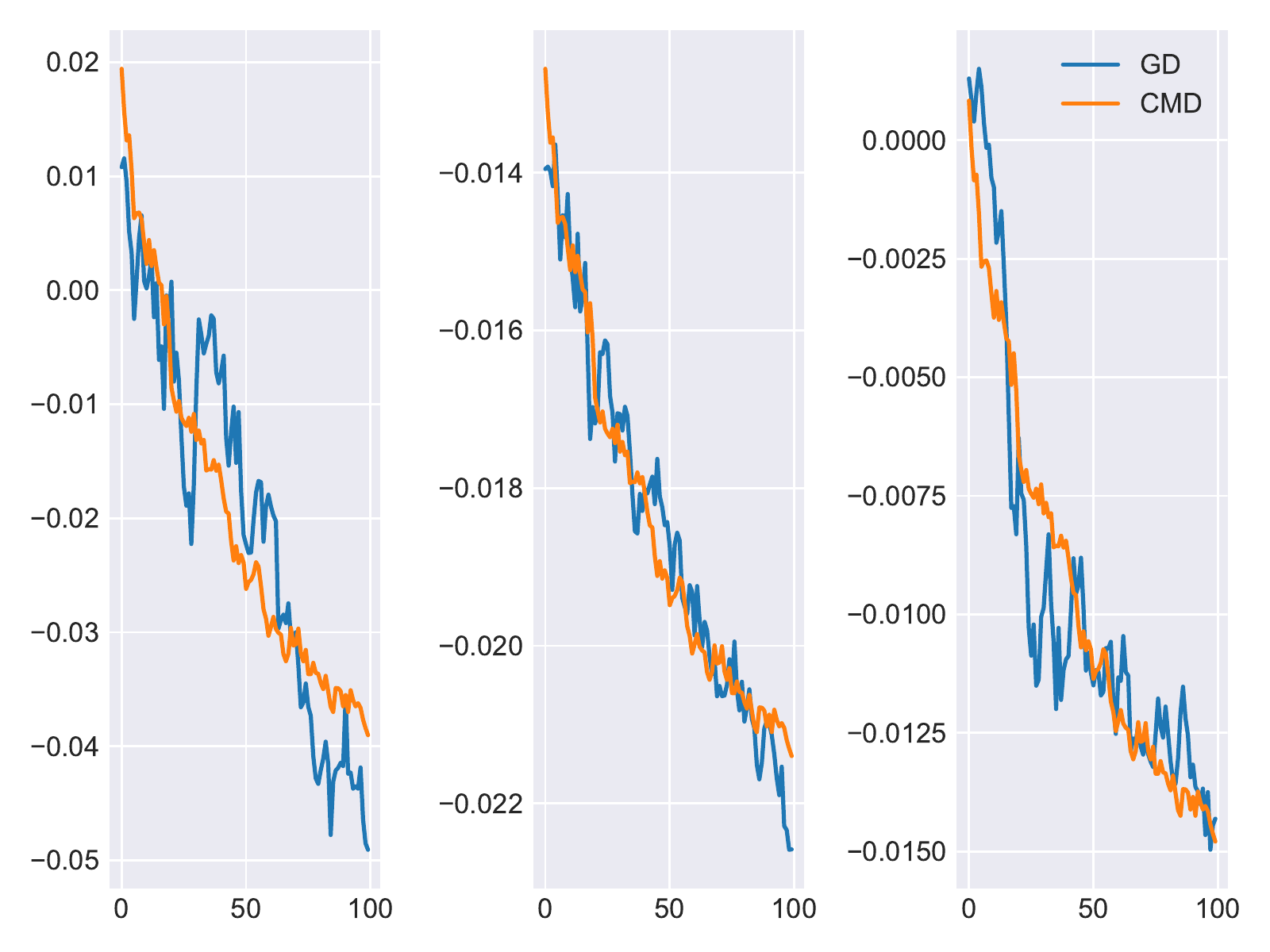}
  \caption{mode 2}
  \label{w_rec:sub3}
  \end{subfigure}
    \begin{subfigure}{.49\textwidth}
  \centering
  \includegraphics[width=\linewidth, height=37mm]{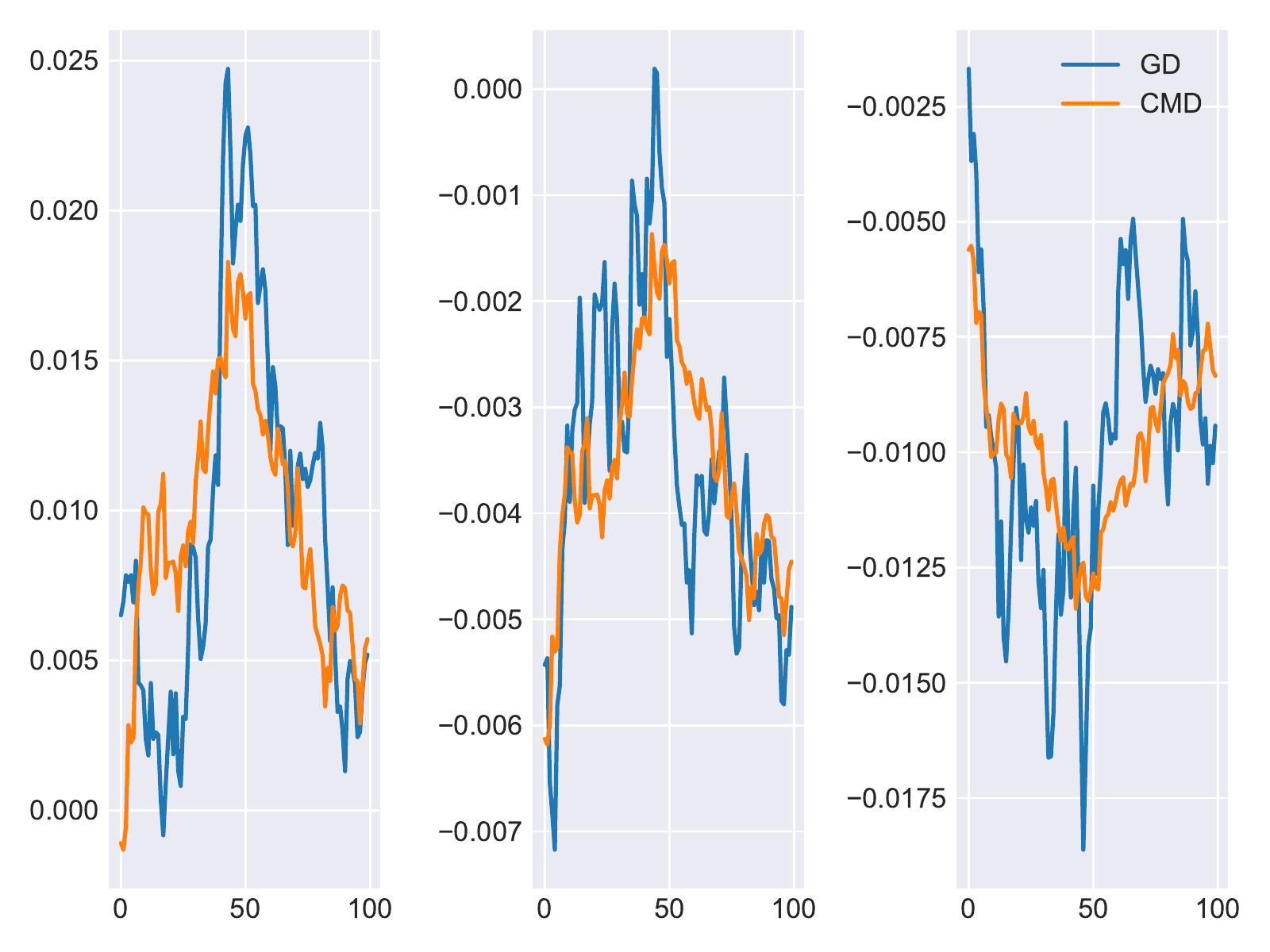}
  \caption{mode 3}
  \label{w_rec:sub4}
  \end{subfigure}
    \begin{subfigure}{.5\textwidth}
  \centering
  \includegraphics[width=\linewidth, height=37mm]{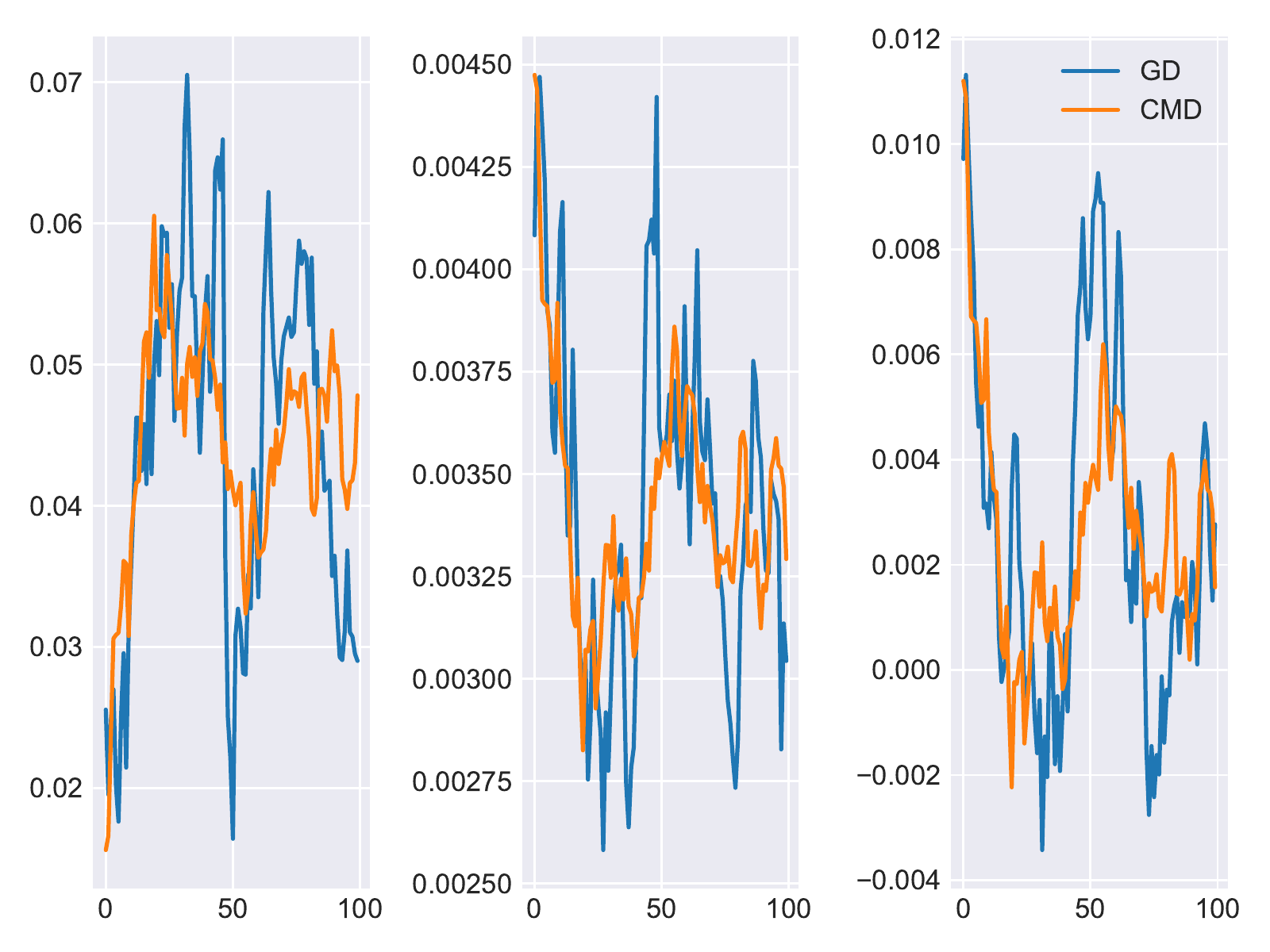}
  \caption{mode 4}
  \label{w_rec:sub5}
  \end{subfigure}
    \begin{subfigure}{.49\textwidth}
  \centering
  \includegraphics[width=\linewidth, height=37mm]{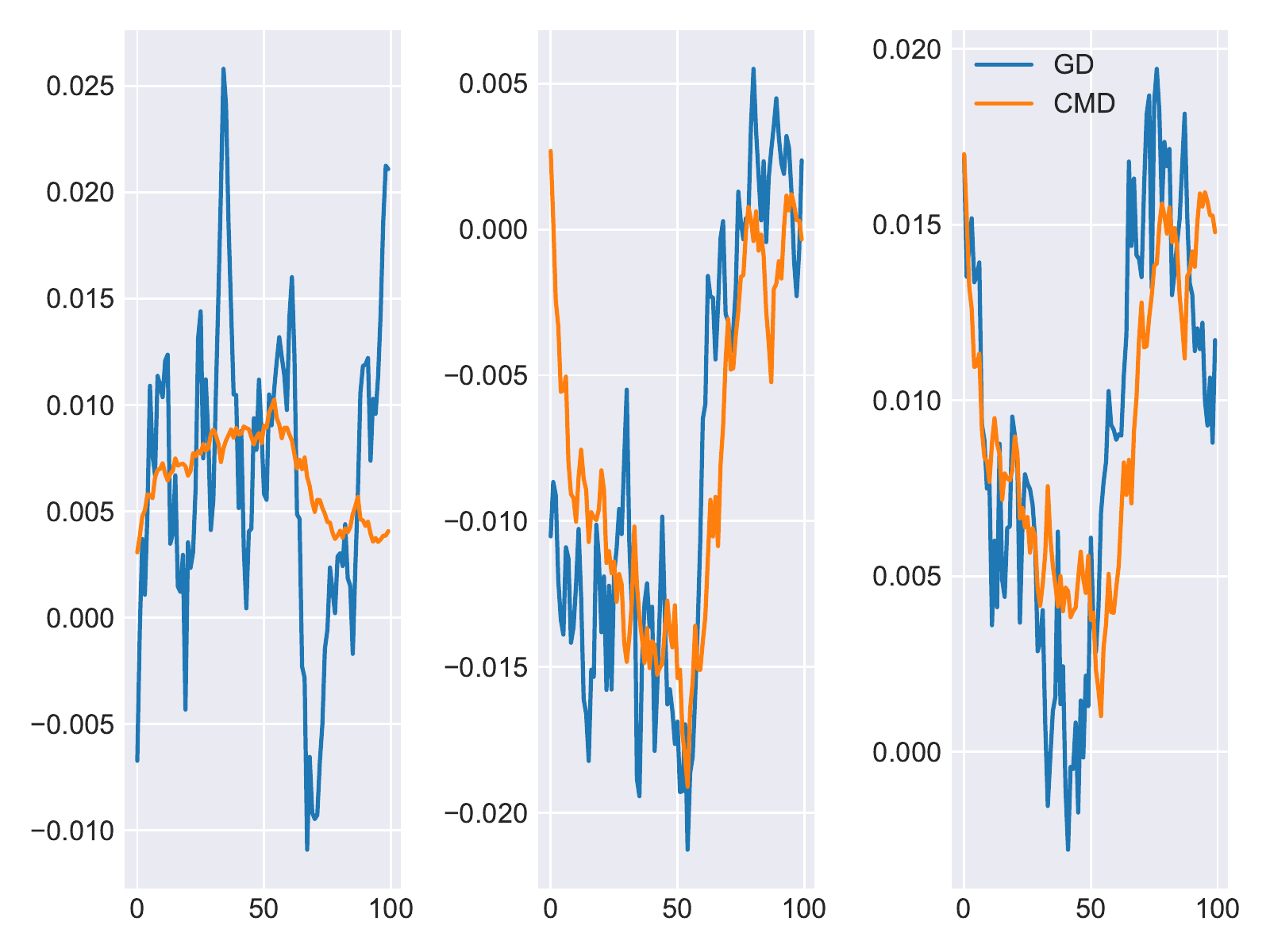}
  \caption{mode 5}
  \label{w_rec:sub6}
  \end{subfigure}
    \begin{subfigure}{.5\textwidth}
  \centering
  \includegraphics[width=\linewidth, height=37mm]{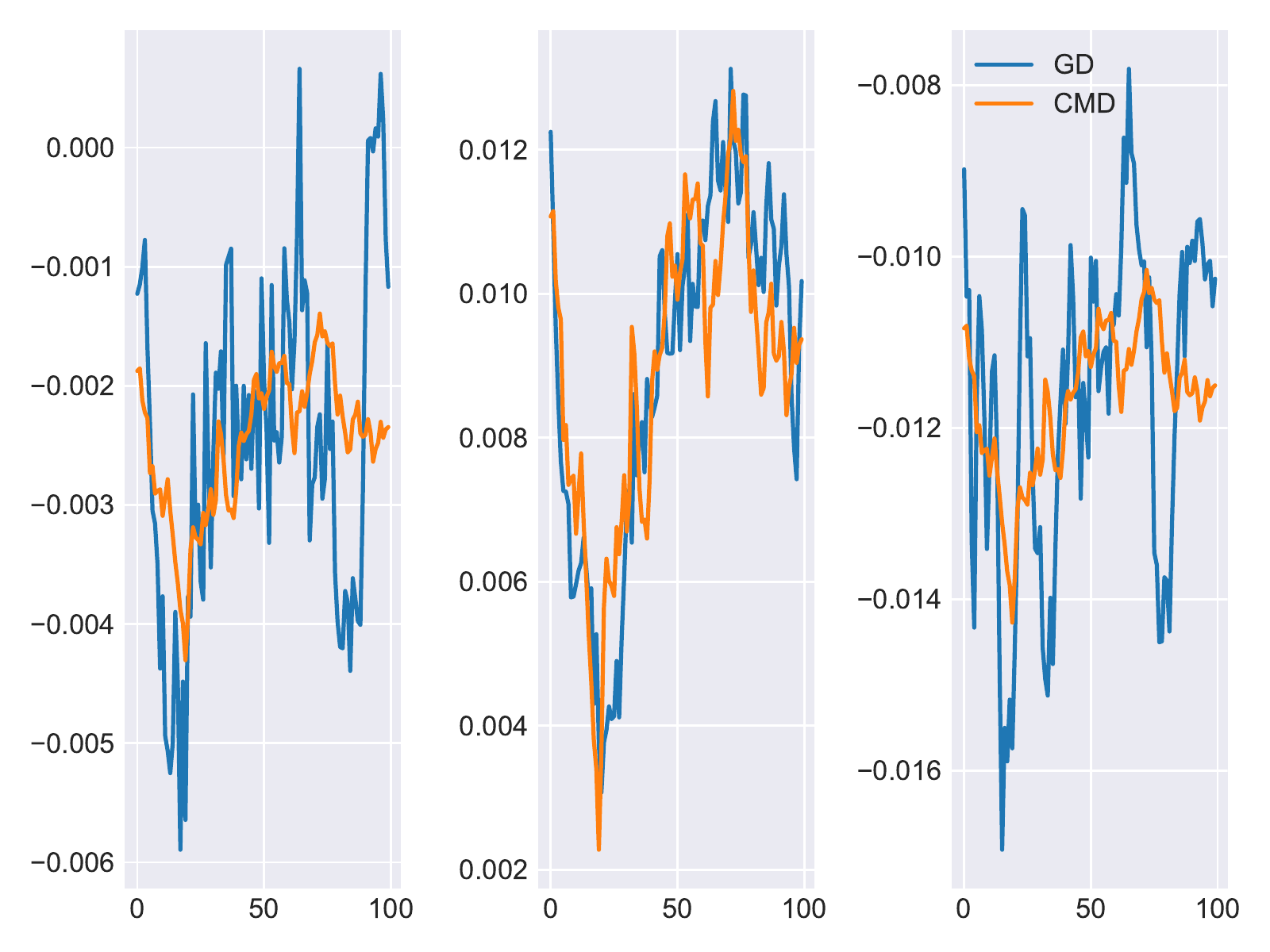}
  \caption{mode 6}
  \label{w_rec:sub7}
  \end{subfigure}
    \begin{subfigure}{.49\textwidth}
  \centering
  \includegraphics[width=\linewidth, height=37mm]{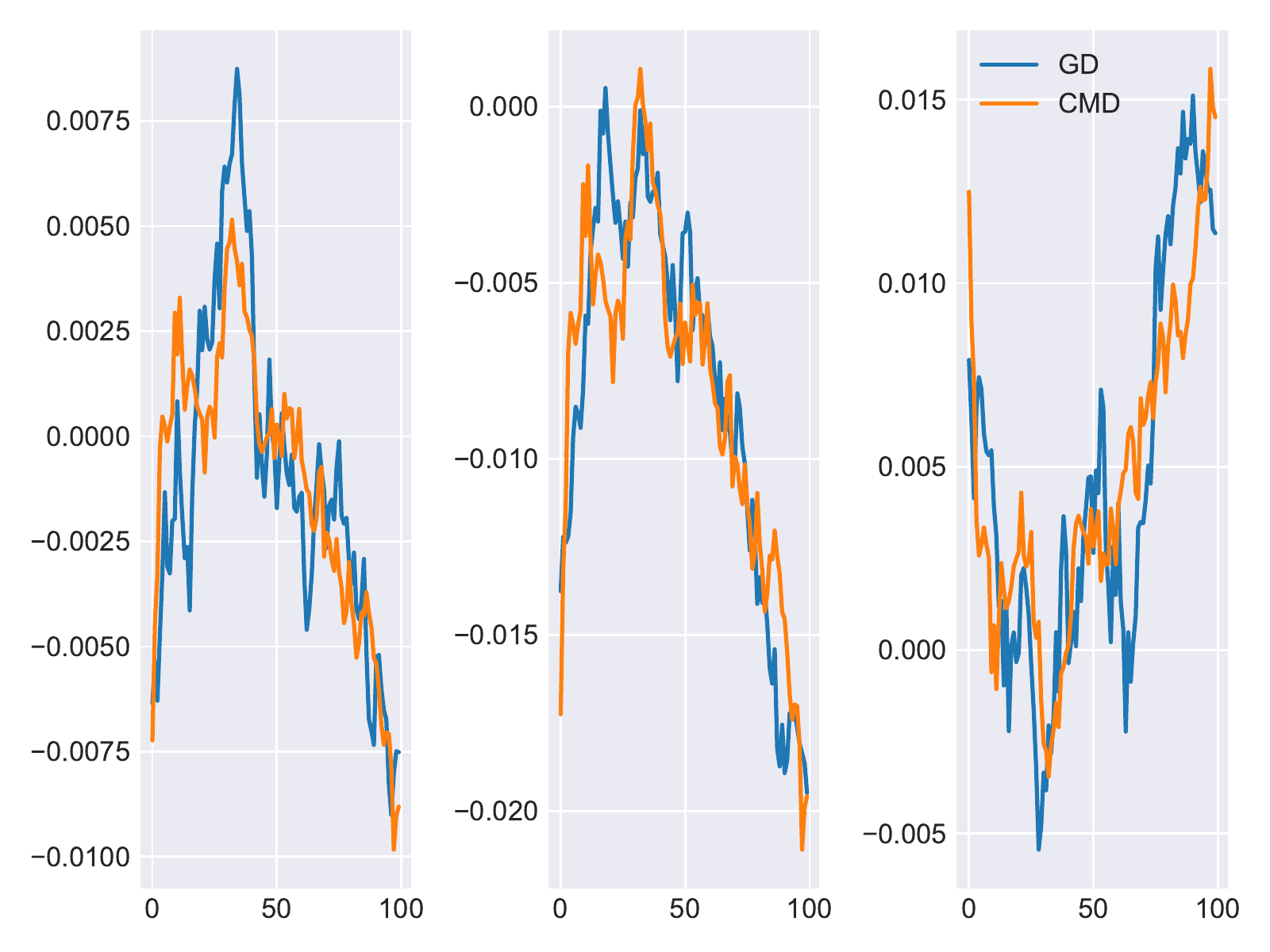}
  \caption{mode 7}
  \label{w_rec:sub8}
  \end{subfigure}
    \begin{subfigure}{.5\textwidth}
  \centering
  \includegraphics[width=\linewidth, height=37mm]{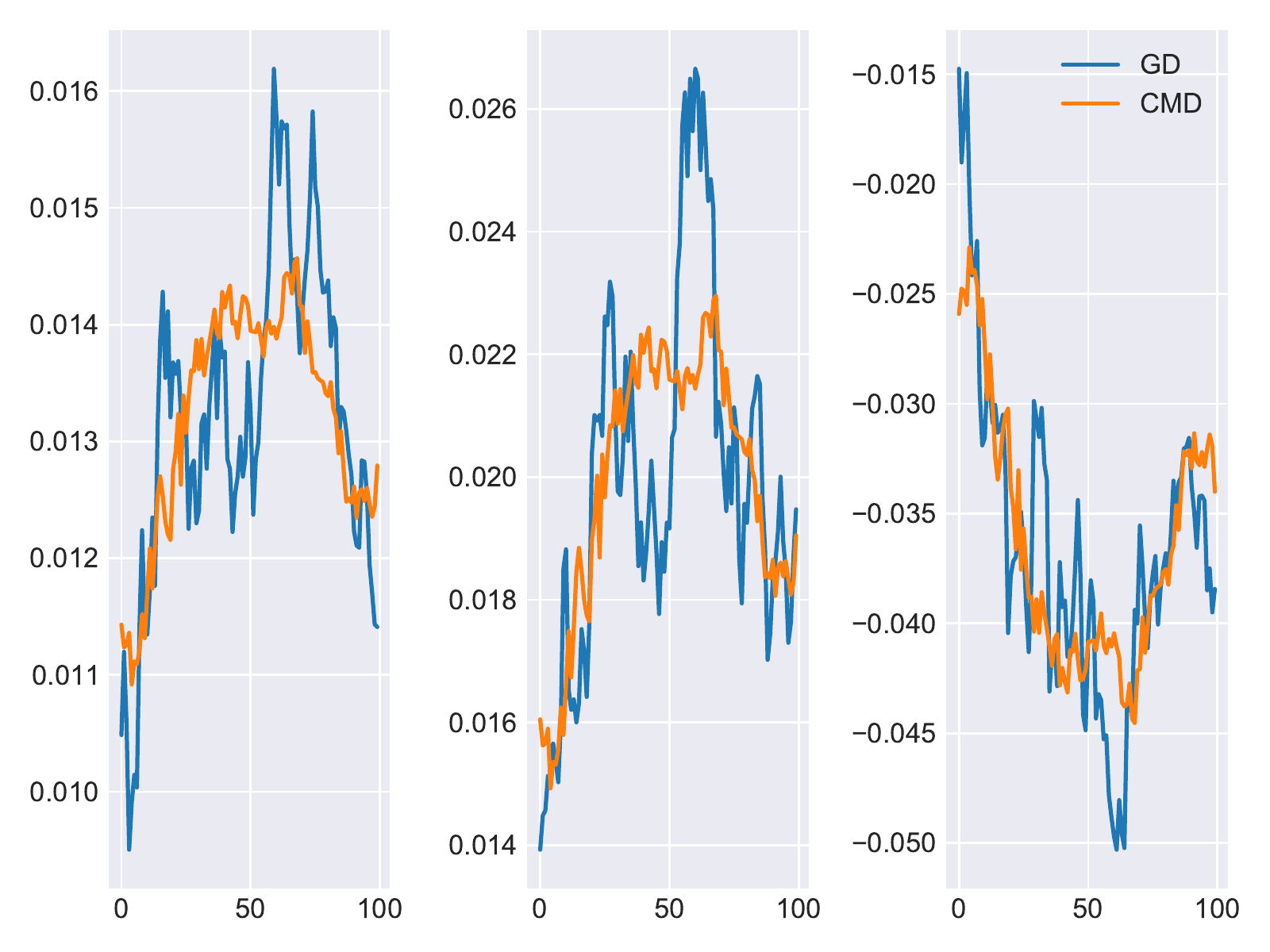}
  \caption{mode 8}
  \label{w_rec:sub9}
  \end{subfigure}
    \begin{subfigure}{.49\textwidth}
  \centering
  \includegraphics[width=\linewidth, height=37mm]{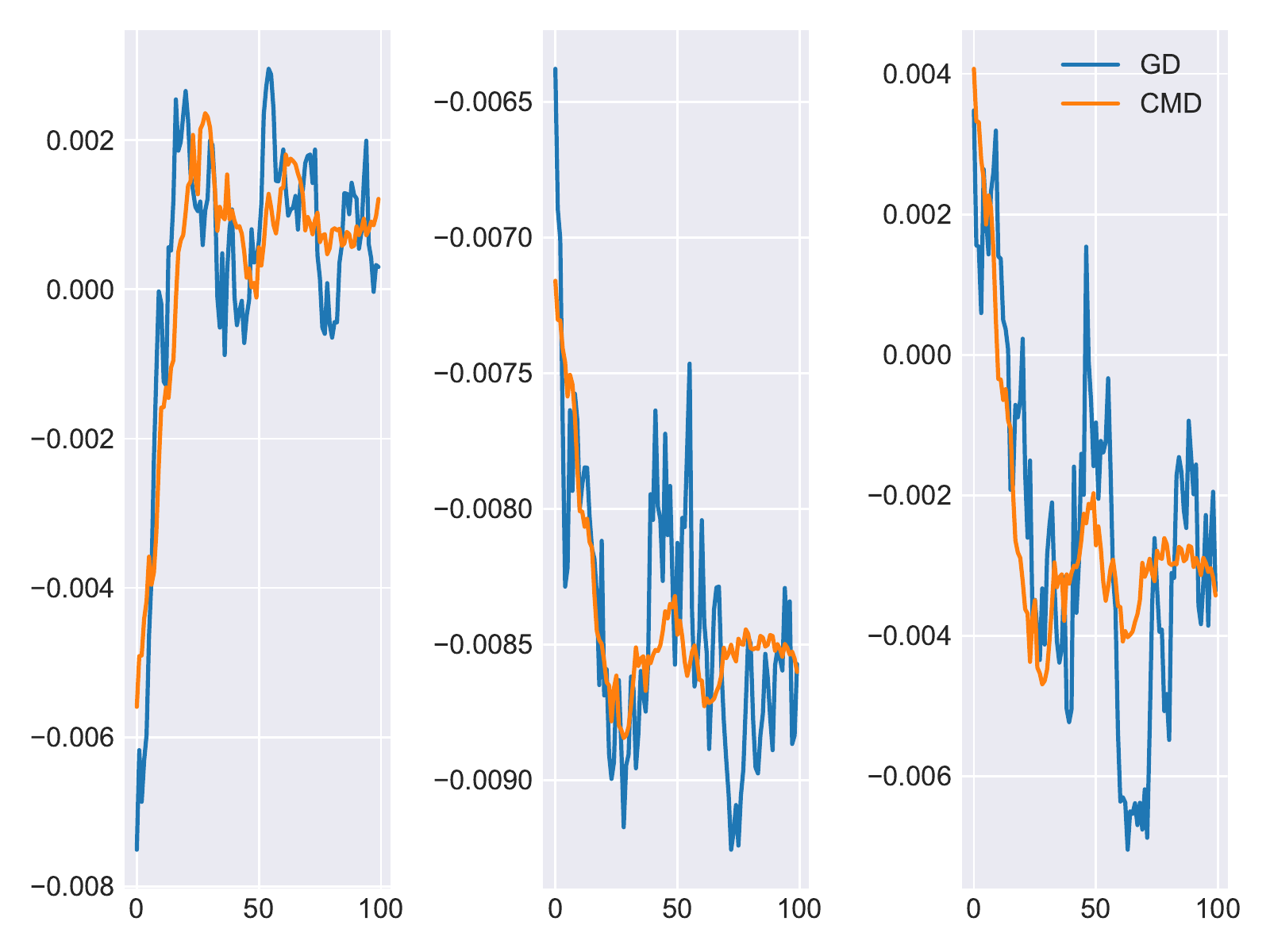}
  \caption{mode 9}
  \label{w_rec:sub10}
  \end{subfigure}
  \caption{3 sampled weights from each mode, and their CMD reconstruction. CIFAR10 classification, Resnet18.
  The CMD modeling provides more stable, less oscillatory solutions.
  }
  \label{w_rec}
\end{figure}

In Fig. \ref{w_rec} we present three weights randomly sampled from each mode, and their CMD reconstruction. These results show that the CMD approximation is much smoother, tends to produce fewer overshoots and less noise. It emphasizes the regularization performed by the algorithm, and explains its stability.
Both figures above are for Resnet18  (CIFAR10).

\begin{figure}[p]
  \centering
\begin{subfigure}{0.7\textwidth}
  \centering
  \includegraphics[width=\linewidth, height=70mm]{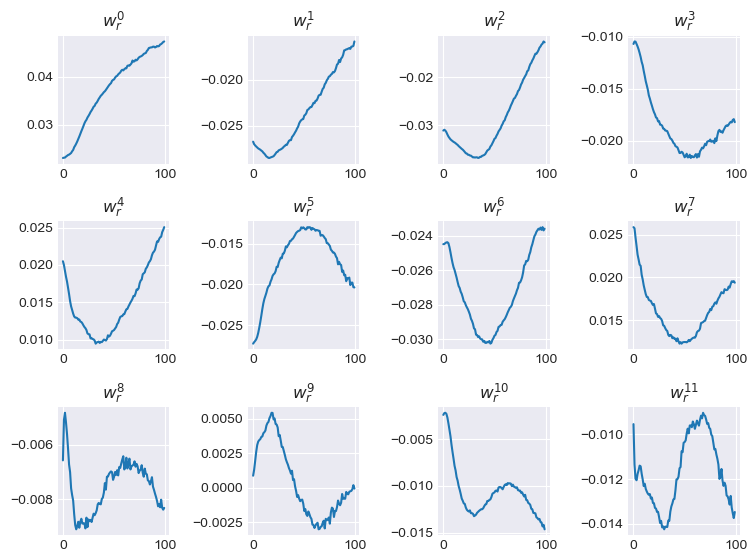}
  \caption{SGD, No Augmentation}
  \label{ref_weights_examination:sub1}
  \end{subfigure}
  \begin{subfigure}{0.7\textwidth}
  \centering
  \includegraphics[width=\linewidth, height=70mm]{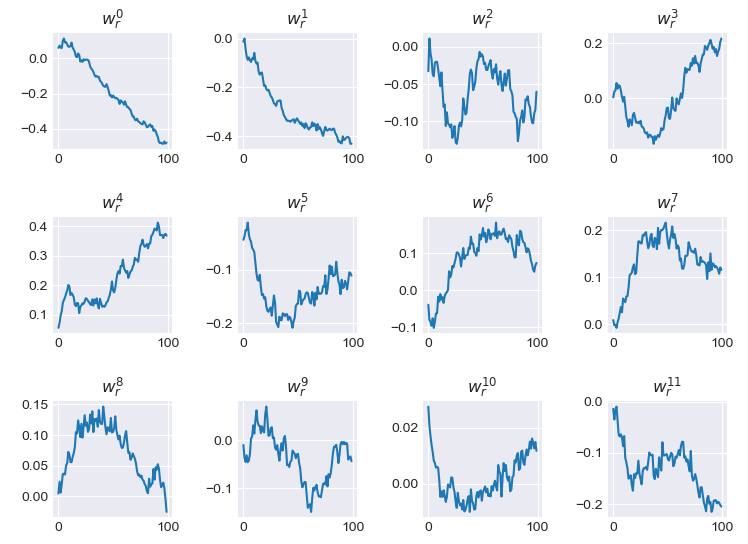}
  \caption{SGD + Augmentation}
  \label{ref_weights_examination:sub2}
  \end{subfigure}
  \caption{Effects of SGD and augmentation on the reference weights. Reference weights of CMD model of SimpleNet (CIFAR10).
  SGD yields the smoothest results, whereas augmentation induces sharp spikes in all modes for the entire training process.
  }
  \label{ref_weights_examination}
\end{figure}

In Fig. \ref{ref_weights_examination} we examine the effect of augmentation on the reference modes. This trial was executed on SimpleNet. 

For better understanding of the dynamics of the training process, we ran this experiment with two configurations: \begin{enumerate}
    \item SGD (batch size=64) with no augmentations.
    \item SGD (batch size=64) with augmentations.
\end{enumerate}

Fig. \ref{ref_weights_examination} presents the difference between the reference weights chosen by the CMD algorithm for each mode in those experiments. One can notice that augmentations indeed influence the weight dynamics in the training process and induce more complex dynamics than the non-augmented case, where the dynamics are much smoother. In addition, this can explain why our model's performance is not sensitive to applying augmentations, since the reference weights of the modes are changed accordingly.


\end{document}